\documentclass[10pt,twocolumn,letterpaper]{article}

\usepackage{cvpr}              

\definecolor{cvprblue}{rgb}{0.21,0.49,0.74}
\usepackage[pagebackref,breaklinks,colorlinks,allcolors=cvprblue]{hyperref}
\usepackage{graphicx}
\usepackage{amsmath}
\usepackage{amssymb}
\usepackage{booktabs}
\usepackage{multirow}
\usepackage{pifont}
\usepackage{xcolor}  
\usepackage{color}
\usepackage{colortbl}
\usepackage{makecell} 
\usepackage[ruled]{algorithm2e} 
\usepackage{xcolor}
\usepackage{tcolorbox}
\usepackage[accsupp]{axessibility}

\newcommand{\greencheck}{{\color{green}\ding{51}}}   
\newcommand{\redx}{{\color{red}\ding{55}}}           

\definecolor{Gray}{gray}{0.90}
\definecolor{white}{rgb}{1.0, 1.0, 1.0}

\definecolor{LightCyan}{RGB}{240, 224, 238}
\newcolumntype{a}{>{\columncolor{LightCyan}}c}

\usepackage{bm}

\usepackage[capitalize]{cleveref}
\crefname{section}{Sec.}{Secs.}
\Crefname{section}{Section}{Sections}
\Crefname{table}{Table}{Tables}
\crefname{table}{Tab.}{Tabs.}

\def\paperID{11196} 
\def\confName{CVPR}
\def\confYear{2025}

\title{STING-BEE \includegraphics[width=0.045\linewidth]{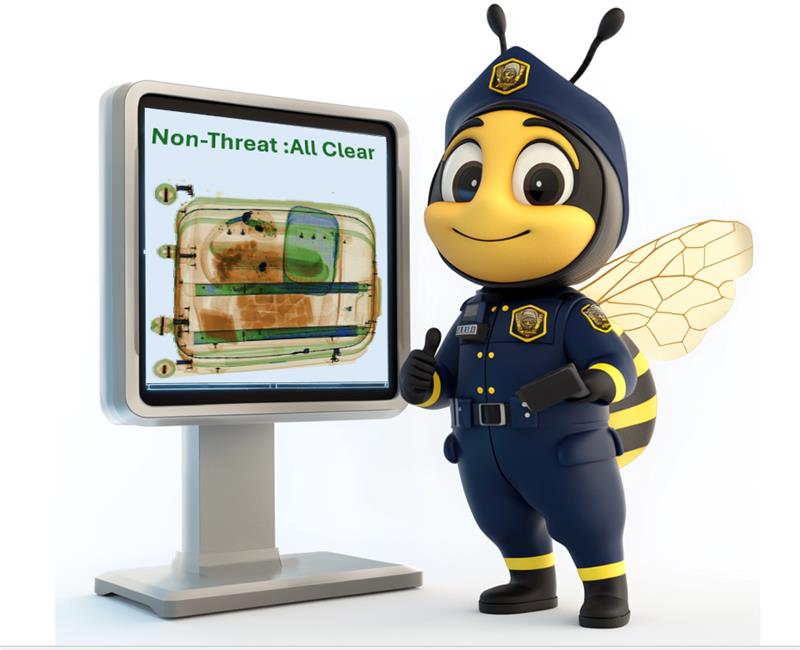}: Towards Vision-Language Model for Real-World X-ray Baggage Security Inspection}

\author{
\begin{tabular}{c}
Divya Velayudhan$^{1}$, Abdelfatah Ahmed$^{1}$\thanks{Equal contribution.}\hspace{2pt}, Mohamad Alansari$^{1}$\footnotemark[1]\hspace{2pt}, Neha Gour$^{1}$, Abderaouf Behouch$^{1}$,\\ Taimur Hassan$^{2}$, Syed Talal Wasim$^{3,4}$, Nabil Maalej$^{1}$, Muzammal Naseer$^{1}$, Juergen Gall$^{3,4}$,\\ Mohammed Bennamoun$^{5}$, Ernesto Damiani$^{1}$, Naoufel Werghi$^{1}$ \\[5pt] 
$^{1}$Khalifa University of Science and Technology \hspace{10pt} 
$^{2}$Abu Dhabi University\\
$^{3}$University of Bonn \hspace{10pt} 
$^{4}$Lamarr Institute for ML and AI \hspace{10pt}
$^{5}$The University of Western Australia
\end{tabular}
}

\begin{document}

\maketitle

\begin{abstract}

Advancements in Computer-Aided Screening (CAS) systems are essential for improving the detection of security threats in X-ray baggage scans. However, current datasets are limited in representing real-world, sophisticated threats and concealment tactics, and existing approaches are constrained by a closed-set paradigm with predefined labels. To address these challenges, we introduce STCray, the first multimodal X-ray baggage security dataset, comprising 46,642 image-caption paired scans across 21 threat categories, generated using an X-ray scanner for airport security. 
STCray is meticulously developed with our specialized protocol that ensures domain-aware, coherent captions, that lead to the multi-modal instruction following data in X-ray baggage security. This allows us to train a domain-aware visual AI assistant named STING-BEE that supports a range of vision-language tasks, including scene comprehension, referring threat localization, visual grounding, and visual question answering (VQA), establishing novel baselines for multi-modal learning in X-ray baggage security. Further, STING-BEE shows state-of-the-art generalization in cross-domain settings. Code, data, and models are available at \small \url{https://divs1159.github.io/STING-BEE/}.

\end{abstract}

\section{Introduction} \label{sec:intro}

\vspace{-1mm}
As global passenger traffic surges, escalating demands on aviation security underscore the importance of meticulous baggage screening to mitigate risks posed by concealed threats while accelerating passenger throughput \cite{WSOL, hixray}. X-ray baggage monitoring is indispensable in security operations, enabling security personnel to examine complex baggage imagery by crosschecking against a set of prohibited items. However, being heavily dependent on human acumen, this procedure is prone to errors due to fatigue, rush-hour distractions, and challenges like overlapping contours, occlusions from high-density objects, high intra-class variance, and pose variations \cite{akcay_survey,survey}. These issues, combined with an evolving threat landscape, further amplify the potential for errors as evidenced by recent studies \cite{skorupski2016human}).

\begin{figure}[!t]
    \centering
    \includegraphics[width=\linewidth]{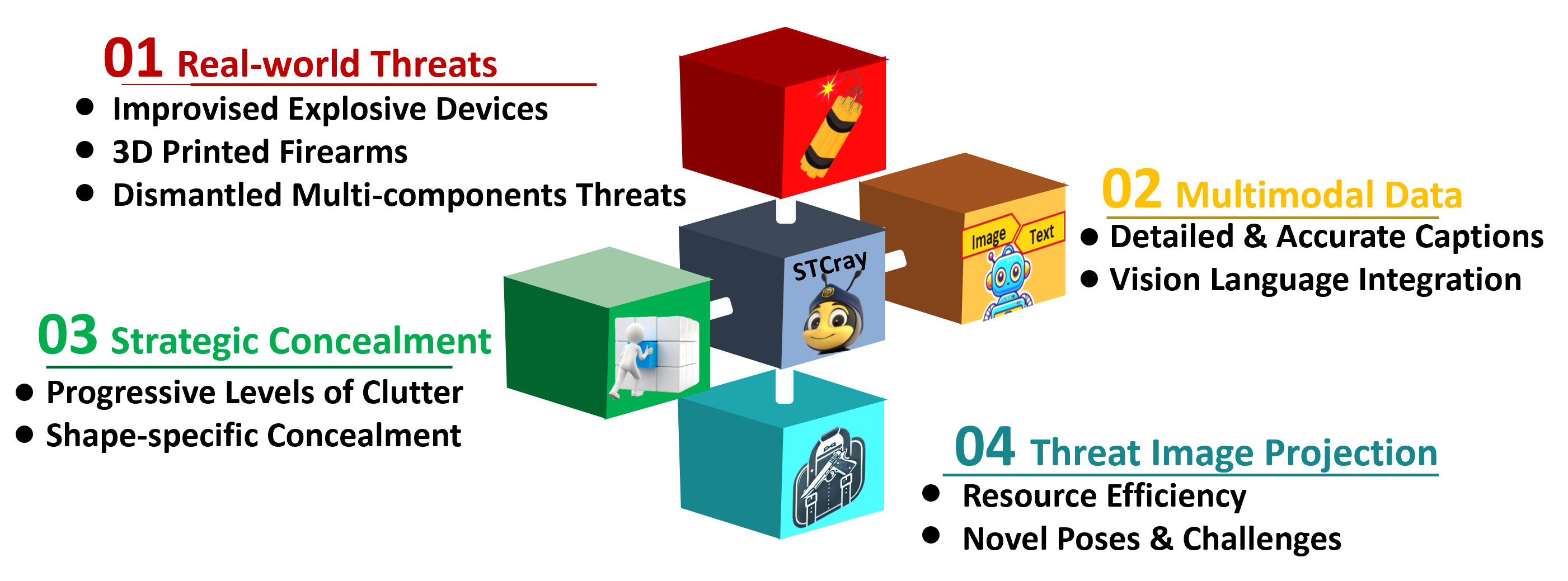}
    
    \vspace{3mm}
    
    \setlength{\tabcolsep}{4pt}
    \scalebox{0.46}[0.46]{
     \begin{tabular}{@{}lccccc@{}}
        \toprule
        \textbf{Dataset} & \textbf{\#Classes} & \textbf{Multimodal} & \makecell{\textbf{Strategic} \\ \textbf{Concealment}} & \makecell{\textbf{Emerging} \\ \textbf{Novel Threats}}  & \makecell{\textbf{Zero-shot} \\ \textbf{Task Capabilities}} \\
        \midrule
        GDXray (JNDE'15) \cite{gdxray}   & 3  & \redx & \redx & \redx &  \redx \\
        SIXray (CVPR'19) \cite{sixray}   & 6  & \redx &  \redx & \redx & \redx \\
        OPIXray (ACMMM'20) \cite{opixray}  & 5  & \redx & \redx & \redx & \redx \\
        HiXray (ICCV'21) \cite{hixray}   & 8  & \redx &  \redx & \redx & \redx \\
        DvXray (IEEE IFS'22) \cite{dvxray}   & 15 &  \redx & \redx & \redx & \redx \\
        CLCXray (IEEE IFS'22) \cite{clcxray}  & 12 &  \redx & \redx & \redx & \redx \\
        PIDRay (IJCV'23) \cite{pidray}  & 15 & \redx &  \redx & \redx & \redx \\
        \midrule
        \textbf{STCray (Ours)}   & 21 & \greencheck & \greencheck & \greencheck & \greencheck \\
        \bottomrule
    \end{tabular}
    }
    \caption{\small (Top) A holistic overview of our STCray, the X-ray dataset with real-world threats and image-text paired data, (Below) A  comparison with public datasets in terms of multi-modality, strategic concealment, emerging novel threats, and zero-shot task.}
    \label{fig_Top_f}
    \vspace{-1.5em}
\end{figure}
\begin{figure*}[t!]
    \centering
    \includegraphics[width=1\linewidth]{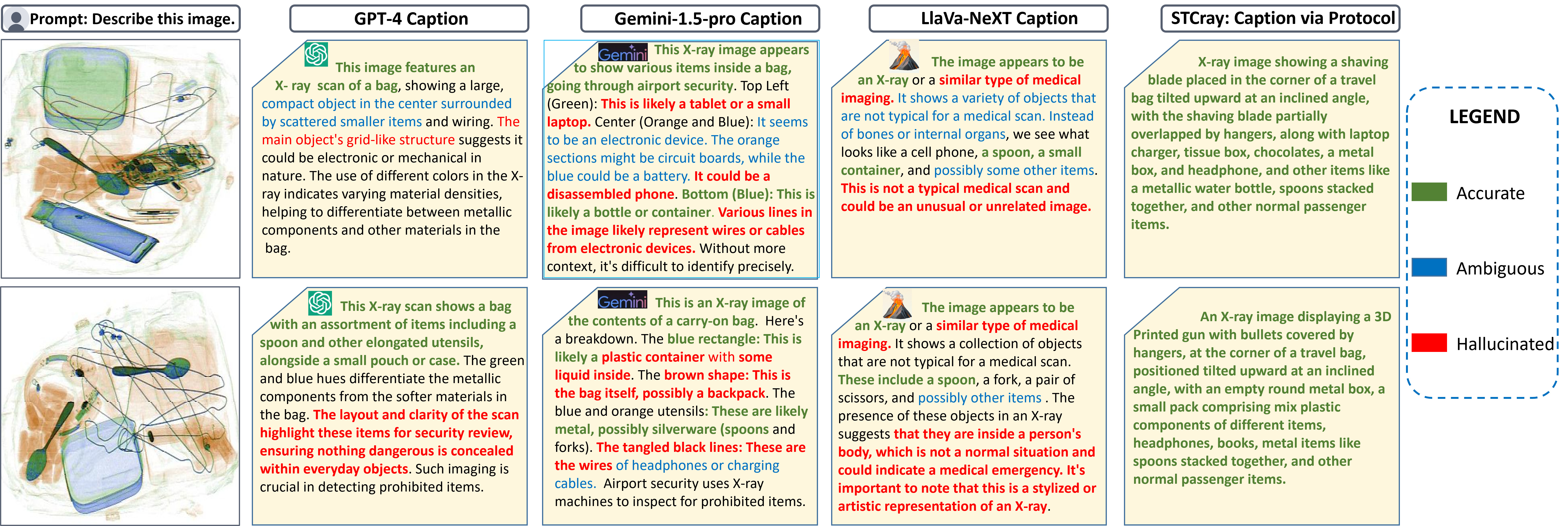} 
    \caption{\small Each row compares GPT-4, Gemini-1.5 Pro, and LlaVa-NeXT captions for the input image from the STCray dataset (first column) with the captions generated by our STING protocol (last column). Parts of the captions highlighted in green are correct, in red are wrong, and in blue are ambiguous. GPT-4 and Gemini-1.5 Pro fail to identify any threat items in both scans e.g., blade (1st-row image) and 3D printed gun (2nd-row image), while LlaVa-Next interprets the baggage X-ray images as medical scans \emph{(best viewed in zoom)}.}
    \label{fig_caption_comp}
    \vspace{-1em}
\end{figure*}

Researchers have leveraged recent strides in deep learning to develop computer-aided baggage screening (CAS) systems to aid in detecting security risks \cite{sixray, opixray, eds_dataset, pidray_iccv}. Current research predominantly relies on the shape and structural cues of the prohibited items via edge \cite{hixray,opixray}, and contour mapping \cite{pixray, hassan2022tensor, BalAffin_Jour} or incorporates attention-based techniques \cite{pidray} and multi-level feature refinement \cite{sixray} approaches to enhance threat detection. These techniques constrain the modeling capabilities of the framework to predefined threat categories. While recent efforts have explored few-shot threat detection by enhancing features for prototype learning, it still remains confined to a closed-set paradigm, unable to generalize beyond the label space of the training set \cite{fsod}. Moreover, existing methods fail miserably when confronted with prohibited items exhibiting significant intra-class variability within the same category due to differences in shape and material properties \cite{recht2019imagenet}. For instance, a model trained using a training set comprising handguns or pistols may fail at inference in detecting rifles due to their structural variations or overlook 3D-printed guns due to their 
unique color palette. Additionally, studies have demonstrated that existing techniques struggle to identify threats across multi-vendor scans, as variations in hardware and imaging mechanisms induce endogenous domain shifts that exacerbate detection failures \cite{eds_dataset}. In summary, current vision-centric approaches \cite{pidray, WSOL} exhibit poor generalization and fail to adequately capture the intricate interrelationships between items in highly cluttered baggage.

A promising solution to these challenges lies in leveraging the capabilities of emerging Vision-Language Models (VLMs) \cite{Clip_radford,gpt4,florence}, which excel at interpreting complex scenes through advanced reasoning and multi-modal learning, facilitating the development of CAS systems that generalize to diverse threat items with minimal customization. However, the primary limitation in applying VLMs to X-ray threat detection stems from the absence of elaborate textual descriptions in existing X-ray security datasets. Existing uni-modal X-ray datasets \cite{sixray, compassxp, pidray_iccv} with simple category labels do not reflect the intricacies of cluttered baggage scans. For example, when threats are deliberately concealed with common items like cables or laptops, mere category labels fail to convey the nuanced context necessary for accurate interpretation. Moreover, unlike medical or web images, where detailed textual descriptions can be obtained from alt-text tagging or expert annotations (by radiologists), generating coherent captions for X-ray security images poses significant challenges \cite{cc12m}. Large multi-modal models (LMMs) like GPT-4 \cite{gpt4} or Gemini \cite{gemini}, primarily trained on natural images, often produce inaccurate or hallucinatory captions when applied to X-ray imagery and often miss the concealed threats (as illustrated in \autoref{fig_caption_comp}). These LMMs lack exposure to the specialized nature of X-ray content, fail to accurately interpret the baggage scan, and produce captions unreliable for effective threat detection. Furthermore, existing datasets do not adequately reflect real-world threats such as IEDs and 3D-printed firearms and concealment scenarios that mimic smuggling tactics — crucial for developing robust Computer-Aided Screening (CAS) systems.

To address these limitations, we introduce the \emph{Strategic Threat Concealment X-ray (STCray)} security inspection dataset, the first multimodal dataset in this domain, meticulously curated using a systematic threat concealment protocol to generate coherent domain-aware captions. STCray comprises 46,642 X-ray images spanning 21 categories with several unique characteristics (\autoref{fig_Top_f}) including: \textbf{a)} Strategic threat concealment protocol with progressively increasing levels of occlusion, facilitating the generation of detailed captions to encapsulate diverse attributes such as threat position and orientation, occluding objects, and degree of occlusion; \textbf{b)} Real-world security threats, notably explosives, and 3D-printed guns.
%
\begin{table*}[!t]
\centering \small
\caption{Comparison of existing X-ray security datasets in terms of the number of X-ray scans, including those containing threats, the number of threat categories, annotations (image labels, bounding boxes, masks, captions), threat material types, application, and public availability. `C', `D', `I', `FD', `DA', and `ZS' represent Classification, Detection, Instance Segmentation, Few-shot Detection, Domain Adaptation, and zero-shot tasks, respectively.}
\label{tab_dataset_comp}
\resizebox{\textwidth}{!}{%
\begin{tabular}{|r|c|cc|cccc|c|c|c|c|}
\hline
\multicolumn{1}{|c|}{\multirow{2}{*}{\textbf{Dataset}}} & \multirow{2}{*}{Classes} & \multicolumn{2}{c|}{Baggage Scans} & \multicolumn{4}{c|}{Annotations} & \multirow{2}{*}{Color} & \multirow{2}{*}{Material} & \multirow{2}{*}{Application} & \multirow{2}{*}{Availability} \\ \cline{3-8}
\multicolumn{1}{|c|}{} &  & Threat & Non-threat & Labels & Bbox & Masks & Captions &  &  &  &  \\ \hline
GDXray (JNDE 2015) \cite{gdxray} & 3 & 8150 & 0 & \ding{51} & \ding{51} & \ding{55} & \ding{55} & Grayscale & 1 & C, D & \ding{51} \\
Dbf6 (ICIP 2017) \cite{akcay2017evaluation}& 6 & 11,627 & 0 & \ding{51} & \ding{51} & \ding{51} & \ding{55} & RGB & 3 & C, D & \ding{55} \\
Compass XP (JXST 2019) \cite{compassxp}& 2 & 258 pairs & 1643 pairs & \ding{51} & \ding{55} & \ding{55} & \ding{55} & \begin{tabular}[c]{@{}c@{}}Grayscale\\  \& RGB\end{tabular} & - & C & \ding{51} \\
SIXray (CVPR 2019) \cite{sixray} & 7 & 8,929 & 1,050,302 & \ding{51} & \ding{51} & \ding{55} & \ding{55} & RGB & 1 & C, D & \ding{51} \\
OPIXray (ACMMM 2020) \cite{opixray}& 5 & 8,885 & 0 & \ding{51} & \ding{51} & \ding{55} & \ding{55} & RGB & 1 & C, D & \ding{51} \\
HiXray (ICCV 2021) \cite{hixray}& 8 & 45,364 & 0 & \ding{51} & \ding{51} & \ding{55} & \ding{55} & RGB & 3 & C, D & \ding{51} \\
DvXray (IEEE IFS 2022) \cite{dvxray} & 16 & 5,496 pairs & 11,000 pairs & \ding{51} & \ding{51} & \ding{55} & \ding{55} & RGB & 4 & C& \ding{51} \\
PIDRay (ICCV 2021/IJCV 2023) \cite{pidray} & 13 & 47,677 & 76,809 & \ding{51} & \ding{51} & \ding{51} & \ding{55} & RGB & 3 & C, D, I & \ding{51} \\
PIXray (IEEE TMM 2022) \cite{pixray}& 15 & 5,046 & 0 & \ding{51} & \ding{51} & \ding{51} & \ding{55} & RGB & 4 & C, D, I & \ding{51} \\
CLCXray (IEEE IFS 2022) \cite{clcxray}& 12 & 9,565 & 0 & \ding{51} & \ding{51} & \ding{55} & \ding{55} & RGB & 4 & C, D & \ding{51} \\
FSOD (ACMM 2022) \cite{fsod} & 20\textsuperscript{*} & 12,333 & 0 & \ding{51} & \ding{51} & \ding{55} & \ding{55} & RGB & 5 & D, F & \ding{51} \\
EDS (CVPR 2022) \cite{eds_dataset}& 10 & 14,219 & 0 & \ding{51} & \ding{51} & \ding{55} & \ding{55} & RGB & 4 & DA & \ding{51} \\
LPIXray (CIPAE 2023) \cite{lpixray}& 18 & 60,950 & 0 & \ding{51} & \ding{51} & \ding{55} & \ding{55} & RGB & 4 & C, D & \ding{55} \\ \hline
\textbf{STCray} & \textbf{21} & 45,693 & 949 & \ding{51} & \ding{51} & \ding{51} & \textbf{\ding{51}} & RGB & 5 & \textbf{C, D, I, ZS} & \ding{51} \\ \hline
\end{tabular}%
}
  \begin{scriptsize}
  \begin{flushleft}
  \centering
    \textsuperscript{*}FSOD \cite{fsod} contributes only 6 categories, while scans corresponding to the remaining 14 categories are taken from OPIXray, HiXray, and EDS datasets.
  \end{flushleft}
  \end{scriptsize}
  \vspace{-1em}
\end{table*}
\begin{figure*}[t!]
    \centering
    \includegraphics[width=1\linewidth]{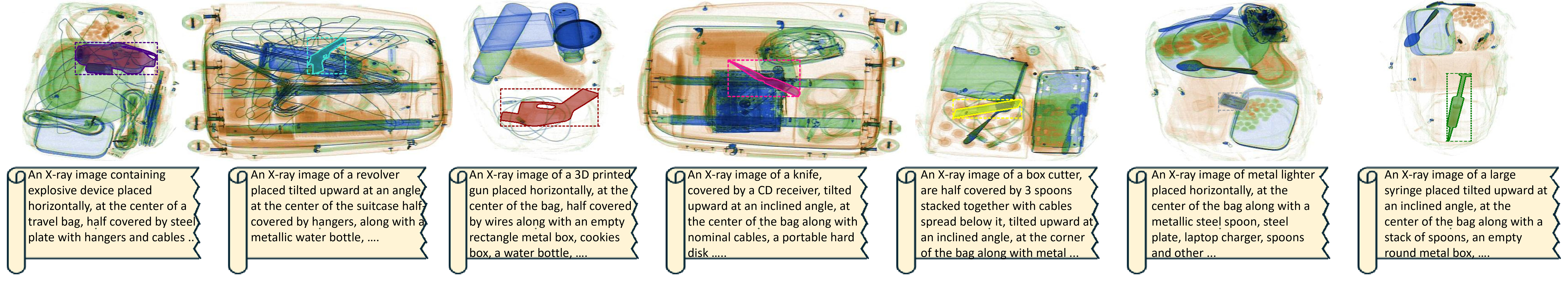} 
    \caption{Sample images from our STCray with instance-level annotations and corresponding captions (\emph{best viewed in zoom}).}
    \label{fig_samples_L8}
    \vspace{-1em}
\end{figure*}
%
Further, to the best of our knowledge, we introduce the first domain-aware visual AI assistant for X-ray baggage screening, named \emph{Strategic Threat Identification aNd Grounding for Baggage Enhanced Evaluation (STING-BEE)}. 
STING-BEE sets new baselines for X-ray baggage security screening, offering 
scene comprehension, referring threat localization, visual grounding, and visual question answering (VQA). 

We summarize our contributions as follows:
\begin{itemize}[noitemsep, left=0pt]
\item \textbf{STCray}: We introduce STCray, the first X-ray baggage security dataset with 46,642 image-caption paired scans spanning 21 categories, including Improvised Explosive Devices (IEDs) and 3D-printed firearms. We meticulously develop STCray by carefully preparing and scanning baggage containing the threat and non-threat items to simulate a realistic environment, following our proposed STING protocol. We spent approximately 3,109 hours creating the STCray dataset, covering baggage preparation, scanning, and annotation (image labels, bounding boxes, masks, captions).
\item \textbf{STING Protocol}: To address the failure of generic VLMs in generating appropriate captions for X-ray baggage imagery, we developed the Strategic Threat ConcealING (STING) 
protocol, which systematically varies the position and angular placement of the threat, progressively increases levels of occlusion. 
This structured approach enables detailed captions (\autoref{fig_samples_L8}) capturing attributes like threat position, orientation, overlap, and occlusion.
\item \textbf{STING-BEE}: Finally, we introduce STING-BEE, the domain-aware visual AI assistant for X-ray baggage screening, trained on the instruction following dataset derived from the image-caption pairs of our proposed STCray dataset. STING-BEE provides a unified platform for scene comprehension, referring threat localization, visual grounding, and VQA, establishing new baselines for X-ray baggage security research. 
\end{itemize}

\noindent We benchmarked STCray dataset using open-source VLMs across diverse tasks, including multi-label classification, threat localization, detection, and VQA. Additionally, we evaluated and compared the cross-domain performance of our STING-BEE 
across diverse X-ray datasets \cite{pidray,sixray,compassxp}. 

\section{Related Work}
\textbf{X-ray Baggage Security Datasets.} Despite the limited availability and sensitive nature of X-ray security inspection data, recent initiatives in constructing X-ray baggage security datasets and public benchmarks~\cite{gdxray,sixray,pidray,opixray,hixray} have led to advancements in computer-aided baggage screening (CAS). GDXray~\cite{gdxray} includes 8,150 grayscale scans with box-level annotations for only 600 images but presents minimal challenges regarding cluttered backgrounds and occlusion. The SIXray dataset~\cite{sixray} is a large-scale benchmark with 1,059,231 scans but has less than 1\% threat images and lacks dense annotations. OPIXray~\cite{opixray} includes 8,885 synthetically generated scans with limited threat category variations, focusing on five types of cutters. Compass-XP~\cite{compassxp} offers paired photographic and X-ray images, comprising 258 threat pairs, but provides only image-level labels, limiting it to classification tasks. HiXray~\cite{hixray}, with 45,364 scans, features eight categories of prohibited items but lacks dense annotations. DvXray~\cite{dvxray} consists of dual-view scans with 5,000 pairs of abnormal scans across 15 threat categories but includes only image-level labels. PIXray~\cite{pixray} is a smaller dataset with 15 categories and dense pixel-level labels. The CLCXray dataset~\cite{clcxray} includes 9,565 scans, focusing exclusively on cutters and liquids. PIDray~\cite{pidray} is a large-scale dataset with 47,677 scans and 12 threat classes, emphasizing intentionally concealed items. Some datasets target specific challenges, such as the EDS dataset~\cite{eds_dataset} with 14,219 scans addressing domain shift, and the FSOD dataset~\cite{fsod} with 12,333 images focused on few-shot learning. Additionally, some studies leverage private datasets~\cite{akcay2017evaluation, lpixray}; for instance, LPIXray~\cite{lpixray}, comprising 60,950 X-ray images across 18 threat categories, includes both real and augmented scans but is not publicly available. \textit{However, existing benchmarks fail to adequately reflect real-world threats and concealment scenarios that mimic smuggling tactics, which are crucial for developing robust Computer-Aided Screening (CAS) systems. Moreover, these benchmarks are uni-modal, lacking the support needed for multi-modal learning.}

 \textbf{Vision-Language Models.} Early vision-language models like CLIP~\cite{Clip_radford} and ALIGN~\cite{align} pioneered the use of large-scale image-text pairs and contrastive learning for zero-shot generalization~\cite{rao2022denseclip, du2022learning, zhang2022pointclip, wasim2023vitaclip, khattak2023promptsrc}. Later, models such as GLIP~\cite{li2021glip} and Grounding DINO~\cite{liu2024groundingdino, wasim2024vgdino} integrated object detection with phrase grounding, while Kosmos-1 and Kosmos-2~\cite{huang2023kosmos-1, peng2023kosmos-2} expanded capabilities to include advanced language understanding and spatial reasoning~\cite{chen2023shikra, you2023ferret}. Latest generation of VLMs enables conversational interactions with visual content by combining a visual backbone~\cite{Clip_radford} with a language model~\cite{vicuna} via cross-modal fusion, like linear projections~\cite{minigpt,Llava} or MLPs~\cite{Llava15}. Models such as LLaVA~\cite{Llava15}, Instruct-BLIP~\cite{dai2023instructblip}, and MiniGPT-4~\cite{minigpt} can perform complex visual reasoning (e.g., in medical \cite{alkhaldi2024minigpt}) and visual grounding (remote sensing~\cite{geochat}) through domain-specific instruction tuning, without any architectural changes. 
\textit{Similarly, there is a need for specialized VLMs for X-ray imagery, as current models—both open and closed source—struggle with the unique challenges of this domain}.

\section{The STCray Dataset} \label{sec:STCray}
\begin{figure*}[!t]
    \centering
    \includegraphics[width=1\linewidth]{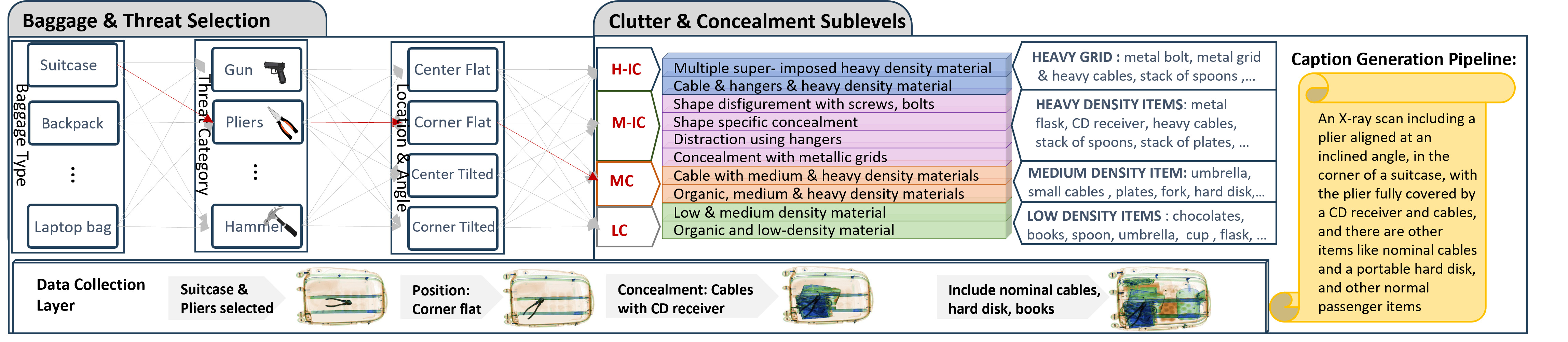} 
    \caption{STING protocol systematically generates captions for X-ray baggage images by selecting the baggage type and threat category (e.g., pliers). It then specifies item location and pose (e.g., corner flat), followed by levels of concealment and clutter, with varying occluding objects and degrees of occlusion and additional normal items. This approach models realistic concealment scenarios, providing detailed information on item location, orientation, and surrounding context to support precise caption generation for each scan.}
    \label{fig_CaptionGen}
\end{figure*}
The STCray dataset features a wide spectrum of threat categories, ranging from sharp metallic threats to flammable materials, including \textit{Explosive}, \textit{Gun}, \textit{3D-printed Gun}, \textit{Knife}, \textit{Cutter}, \textit{Blade}, \textit{Shaving Razor}, \textit{Lighter}, \textit{Syringe}, \textit{Battery}, \textit{Nail Cutter}, \textit{Other Sharp Item}, \textit{Powerbank}, \textit{Scissors}, \textit{Hammer}, \textit{Pliers}, \textit{Wrench}, \textit{Screwdriver}, \textit{Handcuffs}, and \textit{Bullet}. 
\autoref{fig_samples_L8} shows sample images with captions, highlighting security threats with both box- and pixel-level annotations.
Each category features diverse instances varying in shape and material composition, reflecting realistic intra-class variations. For example, the \textit{Lighter} category includes both metallic and non-metallic variants, mirroring real-world security screening diversity.
\textbf{Emerging threats: }Unlike existing datasets that fail to adequately capture real-world threats, STCray emphasizes sophisticated prohibited items, such as IEDs and 3D-printed firearms, alongside realistic concealment scenarios that mimic smuggling tactics. Notably, 3D-printed firearms pose a significant challenge due to their faint outlines, which are difficult to detect in X-ray scans. Explosives are equally challenging due to their non-standard shapes and multi-component structures, including explosive charges enclosed in containers, detonators, activators, and power sources. STCray incorporates both cohesive (compact) and dispersed types within the \textit{Explosive} category, where individual components are either grouped together or spread throughout the baggage and connected by wires, effectively mirroring real-world settings.
\subsection{STING Protocol} \label{sec:STCray_STINGProto}


The STCray data generation follows our systematic STING protocol, which methodically adjusts the position and angle of each threat item within various baggage types—such as suitcases, backpacks, gym bags, and fanny packs—while controlling levels of occlusion. Occlusion levels are progressively increased based on clutter, ranging from minimal to heavy, with intentional concealment strategies. Each level of clutter integrates carefully designed concealment scenarios tailored to the threat’s structure and material, combined with other items of varying material compositions and densities (low, medium, and high). This approach ensures that scans capture both common and sophisticated smuggling tactics, with detailed information on threat location, orientation, and occluding objects (\autoref{supplement: Additional details on STING Protocol}). The STING protocol also enables the generation of precise captions, detailing attributes such as threat position, orientation, material properties, occluding objects, and occlusion level.
\autoref{fig_CaptionGen} illustrates our data collection protocol and caption generation pipeline. The train and test subsets follow this protocol, incorporating diverse baggage types, threat instances, and occluding items in each subset to create a realistic benchmark for evaluating model generalizability.

\begin{figure}[t!]
    \centering
    \resizebox{\columnwidth}{!}{%
    
    \begin{minipage}{0.6\linewidth}
        \includegraphics[width=\linewidth]{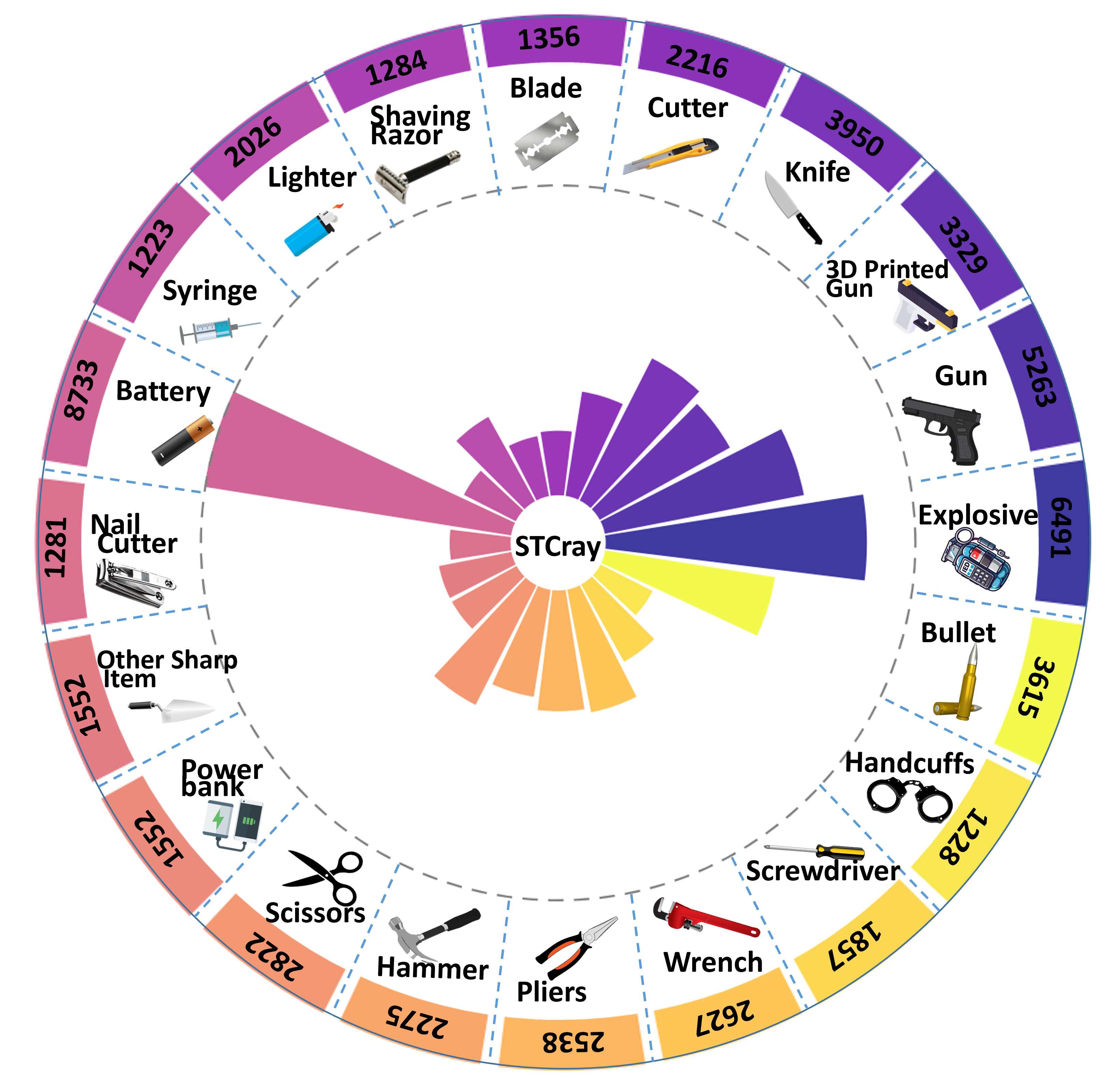}
    \end{minipage}
    \hfill 
    \begin{minipage}{0.4\linewidth}
        \resizebox{0.99\linewidth}{!}{
        \centering
        \begin{tabular}{lrr}
        \toprule
        \multicolumn{3}{c}{\textbf{The STCray DATASET}} \\ 
        \midrule
        Category & \textbf{Train} & \textbf{Test} \\
        \midrule
        Explosive & 2758 & 3733 \\
        Gun & 4702 & 561 \\
        3D Printed Gun & 2125 &	1204 \\
        Knife & 3386	&564\\
        Cutter & 1629	&587 \\
        Blade & 904 &	452 \\
        Shaving Razor & 873 &	411 \\
        Lighter & 840 &	1186 \\
        Syringe & 854	& 369 \\
        Battery & 4091 &	4642 \\
        Nail Cutter & 879	&402 \\
        Other Sharp Items & 880	&672 \\
        Powerbank & 1037 &	515 \\
        Scissors & 1348	& 1474 \\
        Hammer & 1606 &	669 \\
        Pliers & 1200	& 1338 \\
        Wrench & 2121	& 506 \\
        Screwdriver & 1340	& 517\\
        Handcuffs &863	&365\\
        Bullet & 3051 &	564 \\
        \midrule
        \textbf{Total} & \textbf{36,487} & \textbf{20,731} \\ 
        \bottomrule
    \end{tabular}
        }
    \end{minipage}
    } 
    \caption{Instance-wise distribution of threat categories in the STCray dataset. Left: Radial plot depicting overall counts; Right: Table summary across train and test sets.}
    \label{fig_Instance_Data_dist}
    \vspace{-1.5em}
\end{figure}

\textbf{Statistics:} 
Our STCray dataset is comprehensively annotated, featuring 46,642 samples with detailed descriptions, bounding boxes, and pixel-level labels. The dataset comprises 30,044 training and 16,598 test images, ensuring a robust evaluation set. It includes 57,218 threat instances across 20 threat categories, with instance-wise distributions shown in \autoref{fig_Instance_Data_dist}. Notably, it contains 36,438 single-threat images and 9,255 scans with multiple threats (\autoref{tab_single_multiple_threatImages}).
\begin{table}[!t]
    \centering
    \caption{Statistical distribution of images containing single and multiple threat instances in our dataset}
    \label{tab_single_multiple_threatImages}
    \resizebox{\columnwidth}{!}{%
        \begin{tabular}{lcc}
        \toprule
        \textbf{Mode} & \textbf{Single-Instance Threat Images} & \textbf{Multi-Instance Threat Images} \\
        \midrule
        Train & 23,622 & 5,822 \\
        Test & 12,816 & 3,433 \\
        \midrule
        \textbf{Total} & \textbf{36,438} & \textbf{9,255} \\
        \bottomrule
        \end{tabular}
    }
    \vspace{-1.5em}
\end{table}

\section{STING-BEE} \label{sec:stingbee}
\subsection{Architecture}

\begin{figure*}[!t]
    \centering \small
    \includegraphics[width=\linewidth]{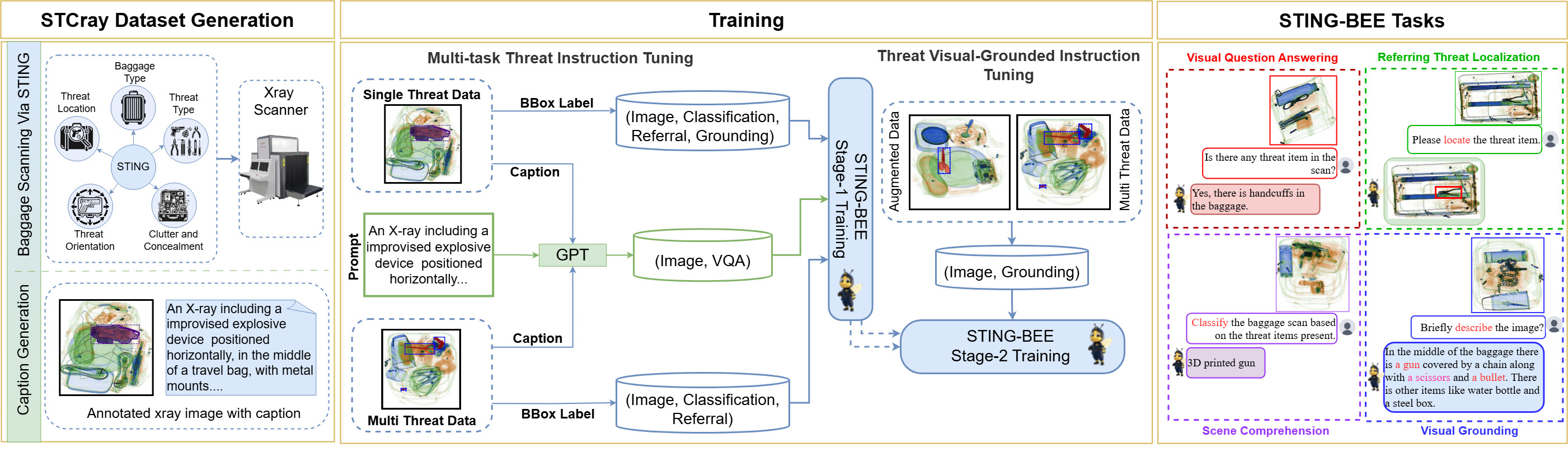} 
    \caption{STING-BEE Training and Evaluation Pipeline: (\emph{Left}) STCray Dataset Collection, capturing X-ray images with systematic variations in threat type, location, and occlusion, along with detailed captions and bounding box annotations; (\emph{Center}) Multi-modal Instruction Tuning, consisting of Multi-task Threat Instruction Tuning and Threat Visual-Grounded Instruction Tuning (\emph{Right}) Examples of STING-BEE evaluation tasks including Scene Comprehension, Referring Expression, Visual Grounding, and VQA.}
    \label{fig_StingBee_arch}
    \vspace{-1.5em}
\end{figure*}


STING-BEE builds upon the LLaVA architecture \cite{Llava15} and consists of three main components: an image encoder \( \bm{E} \), a two-layer cross-modal projector \( \bm{P} \), and a large language model (LLM) \cite{vicuna}. Unlike LLaVA, which is limited to visual chat capabilities, STING-BEE uses specialized tokens to effectively distinguish between various vision-language tasks and deliver responses with tailored granularity—from providing X-ray scan descriptions to pinpointing threat locations based on the user query (\autoref{fig_StingBee_arch}.{\small STING-BEE Tasks}).



The image encoder \(\bm{E}\), a CLIP-pretrained Vision Transformer (ViT-L/14) \cite{Clip_radford}, extracts features from the input scan  \(\bm{x}_{i}\) , resulting in a feature sequence  \(\bm{f}_{i} = \bm{E}(\bm{x}_{i}) \in \mathbb{R}^{m \times n} \) . A trainable two-layer MLP projector  \(\bm{P}\)  with GeLU activation then aligns these visual features  \(\bm{f}_{i}\) with large language model, producing k-dimensional output. The transformation is formally expressed as:
\[
\bm{f}_{i}' =  \bm{MLP}_{p} \cdot \bm{f}_{i},
\]
where \( \bm{MLP}_{p} \) is the learned weights of \( \bm{P} \), and \( \bm{f}_{i}' \) denotes the aligned visual tokens.
%
%
Finally, the large language model processes both the projected visual tokens \( \bm{f}_{i}' \) and tokenized language queries to generate responses tailored to specific language-based vision tasks e.g., referring threat localization (\autoref{fig_StingBee_arch}.{\small STING-BEE Tasks}).
\textbf{Task Identification Tokens:} 
STING-BEE is designed for interpreting complex X-ray baggage imagery, unifying vision-language tasks like scene comprehension, referring threat localization, visual grounding, and VQA. Simply aligning visual tokens with large language models does not adequately differentiate between tasks, particularly when responses demand different levels of spatial awareness.
%
To address this, we introduce specialized task-identification tokens \cite{minigptv2, llava_med, geochat}: a \texttt{\small[refer]} token for referring threat localization and a \texttt{\small[grounding]} token for visual grounding. When appended with a user query, these tokens enable the model to easily differentiate between different tasks (\autoref{tab_InstructioData_format}). Scene comprehension and VQA tasks do not require such tokens, as they involve only image-level context and textual responses without spatial references.

\subsection{Multi-task Instruction Tuning}
\label{subsec: Multi-task Instruction Tuning}
%
To train, STING-BEE for X-ray security screening, we created an instruction-following dataset by creating question-answer pairs for scans in our STCray dataset. These question-answer pairs follow structured templates specific to the language-based vision tasks (\autoref{tab_InstructioData_format}).
%
\begin{table}[t!]
\centering
\caption{Task-specific formats for X-ray security Instruction data.}
\label{tab_InstructioData_format}
\setlength{\tabcolsep}{2pt}
    \scalebox{0.6}[0.6]{
\begin{tabular}{|l|c|}
\hline
\textbf{Task} & \textbf{Query Formats for Instruction following Data} \\ \hline
Scene comprehension & 
    \begin{tabular}[c]{@{}l@{}}
        Classify the baggage scan into one or more categories based on the\\ threats present, if any, or classify it as nonthreat. \\
        Focusing on prohibited items present in the image, classify them into the\\ following categories. If there is no threat, classify it as nonthreat.
    \end{tabular} \\ \hline
Referring Expression &
    \begin{tabular}[c]{@{}l@{}}
        {[refer]} Give the location of \textless p\textgreater{} Threat Category\textless /p\textgreater \\ 
        {[refer]} Please find the \textless p\textgreater{} Threat Category\textless /p\textgreater
    \end{tabular} \\ \hline
Visual grounding & {[grounding]} Describe the baggage scan, focusing on the threats if any. \\ \hline
\end{tabular}
}
\vspace{-2em}
\end{table}
\textbf{Scene Comprehension.} This task requires STING-BEE to process the baggage scan \( \bm{x} \) along with a user query \( \bm{q} \) to generate single or multiple words (e.g., “gun, knife”) identifying the concealed threats within cluttered baggage. We generated a total of 30,044 instructions in this category for the STCray training set. Each instruction is based on predefined questions, with answers derived directly from the STCray dataset labels. For example, a question prompt may be: \textit{“Classify the image based on the presence of the following threat classes: explosive, gun, 3D printed gun, knife, cutter, blade, shaving razor, lighter, syringe, battery, nail cutter, other sharp items, powerbank, scissors, hammer, pliers, wrench, screwdriver, handcuffs, bullet, nonthreat. If no threats are present, classify the image as nonthreat.”}.
\textbf{Referring Threat Localization.} This task requires identifying the location of prohibited items within obscured X-ray baggage scans based on user query. To this end, we generated 36,487 referring expression instructions. When presented with an input scan \( \bm{x} \) and a query, such as “Give the location of \texttt{<p>}gun\texttt{</p>},” and using the \texttt{\small[refer]} token, STING-BEE generates a response that includes bounding box coordinates, expressed in a textual format as \( B = \{ b_{x_{\text{min}}}, b_{y_{\text{min}}}, b_{x_{\text{max}}}, b_{y_{\text{max}}} \} \), where \( (b_{x_{\text{min}}}, b_{y_{\text{min}}}) \) denote the top-left corner, and \( (b_{x_{\text{max}}}, b_{y_{\text{max}}}) \) denote the bottom-right corner of the bounding box, normalized to the [0,100] range. We generated instructions based on predefined questions, with answers mapped to normalized bounding box annotations for the queried threat items from the STCray dataset. The \texttt{\small[refer]} token marks referring expression tasks, enabling STING-BEE to deliver precise spatial information for each queried item. 
\textbf{Visual Grounding.} In visual grounding, STING-BEE generates detailed descriptions of baggage scans, identifying threats and their spatial locations using bounding box coordinates. We created 29,444 grounding instructions through question-answer pairs, generated by sampling from a predefined set of questions. The answers were derived from captions produced using the STING protocol. We enhanced these captions to follow a consistent response format, incorporating normalized bounding box coordinates to precisely indicate object positions. The \texttt{\small[grounding]} token was used to distinguish grounding tasks, prompting STING-BEE to provide interleaved responses that incorporate both descriptive text and spatial localization. For example, given an image-caption pair with the caption \textit{"In the middle of the baggage scan, there is a plier covered by hangers and cables,"} we generate a question-answer pair such as: \textit{"[grounding] Describe the baggage scan focusing on threats if present."}
     \textit{"In the middle of the baggage scan, there is \texttt{\small<p>a plier</p>} \texttt{\small\{<31><31><63><42>\}} covered by hangers and cables."}
\textbf{VQA.} This task involves multi-round queries addressing various aspects of the input baggage scan, such as threat identification, threat position, occlusion, and complex reasoning. VQA tests both image-level and object-specific comprehension, enabling the model to handle intricate queries within the domain. We generated 30,044 visual question-answer conversations using the STCray captions, with GPT-4 \cite{gpt4} providing diverse and context-rich responses (see \autoref{supplement: Instruction following Dataset} for the details).

\begin{table*}[t!]
    \centering
    \caption{Visual Question Answering (VQA) performance across seven question categories from in-domain and cross-domain datasets.}
    \vspace{-1em}
    \setlength{\tabcolsep}{4pt}
    \scalebox{0.75}[0.75]{
    \begin{tabular}{@{}lccccccccc@{}}
        \toprule
        \rowcolor{Gray}
        Model & Instance Location & Complex Reasoning & Instance Identity & Instance Counting & Misleading & Instance Attribute & Instance Interaction & Overall \\
        \midrule
        Florence-2 \cite{florenceV2} 
            & 30.11 & 37.50 & 39.84 & 29.95 & 21.16 & 35.80 & 29.12 & 32.27 \\
        MiniGPT \cite{minigptv2} 
            & 43.22 & 60.21 & 54.79 & 28.16 & 19.64 & 32.85 & 29.74 & 36.62 \\
        LLaVA 1.5 \cite{Llava15}  
            & 29.73 & 56.67 & 74.04 & 34.24 & 13.76 & 40.85 & 24.03 & 41.94 \\
            \midrule
            \rowcolor{LightCyan}
        STING-BEE (Ours) 
            & \textbf{49.22} & \textbf{79.21} & \textbf{80.04} & \textbf{45.24} & \textbf{27.76} & \textbf{52.85} & \textbf{35.03} & \textbf{52.81} \\
        \bottomrule
    \end{tabular}%
    }
    \label{tab_vqa_results}
    \vspace{-1em}
\end{table*}

\subsection{Training Strategy}
We train STING-BEE in a multi-stage process to build domain expertise and grounding capabilities tailored for complex X-ray baggage analysis (\autoref{fig_StingBee_arch}.{\small Training}).
This training leveraged Low-Rank Adaptation (LoRA) to tune the LLM while preserving its general language understanding. Additionally, the MLP projector was also trained \cite{minigptv2, geochat}. 
\emph{Multi-task Threat Instruction Tuning:} we first align STING-BEE on a subset of 120,190 instruction training pairs (\autoref{subsec: Multi-task Instruction Tuning}) of single and multi-threat scans for scene comprehension, referring expression, and VQA excluding the visual grounding instruction.  This allows the model gain generic understanding about X-ray data. 
\emph{Threat Visual-Grounded Instruction Tuning:} We focus on spatial reasoning and grounding abilities in the second stage by further tuning the model with grounding instructions specific to multi-threat images in STCray, supplemented with approximately 25,000 CT-2-Xray-augmented images to further enhance data diversity. This realistic augmentation technique leverages CT volumes of threat items to generate 2D projections with varied X-ray attenuation levels and rotations, creating synthetic scenarios that simulate diverse orientations
(see \autoref{supplement: CT-2-X-ray Augmentations} for details). This stage provided an additional 31,000 grounding-specific instructions, enhancing STING-BEE's fine-grained spatial understanding.

\section{Experiments}

\vspace{-1mm}
\textbf{Implementation.} We initialize our STING-BEE model with a CLIP-pretrained ViT-L/14 as the visual encoder, the LLaVA 1.5 multi-modal projector, and the Vicuna-v1.5 \cite{chiang2023vicuna} LLM. We employ LoRA fine-tuning with a designated rank \( r = 64 \), targeting the query (\( W_q \)) and value (\( W_v \)) projection matrices of each attention layer in the LLM \cite{minigptv2, geochat}. We use AdamW optimizer at a learning rate of \(2 \times 10^{-5}\), using a cosine learning rate scheduler for smooth convergence.
We fine-tune our model for one epoch with a global batch size of 96 followed by the second stage tuning for one epoch on grounding-specific instructions using two A100 GPUs.

\subsection{Scene Comprehension}
\textbf{Evaluation Dataset:} We assess scene comprehension capabilities, in a \textbf{cross-domain} setting on an extensive test set comprised of 13,412 scans from SIXray \cite{sixray} across five threat categories, 47,000 from PIDray \cite{pidray} with 12 classes, and 1900 from COMPASS-XP \cite{compassxp} with two classes. Each dataset presents unique scanning characteristics due to differing X-ray scanners and threat categories, introducing domain shifts that test the robustness of our model. We prompt the models to classify the input scan to identify the multiple threat categories: \emph{''Classify the scan into exactly one or more categories. Categories: baton, pliers, hammer, ... Answer only these categories nd no lengthy sentences.''} 
\textbf{Results:} STING-BEE outperformed state-of-the-art VLMs across both F1 Score and mAP metrics (\autoref{tab_zeroshot_Scenecomprehension}). Particularly, STING BEE achieved an F1 Score of 34.69 and mAP of 29.80, significantly surpassing the model, MiniGPT-v2, which obtained an F1 Score of 18.45. 
The result highlights the ability of STING-BEE trained on our proposed dataset, to adapt to \textbf{unseen} diverse threat categories and domain shifts, maintaining high performance despite variations in X-ray scanners and image contexts.
\begin{table}[t!]
    \centering \small
    \caption{\textbf{Cross-domain} Scene Comprehension Performance. STING BEE (Proposed) outperforms other SOTA VLMs.}
    \vspace{-1em}
    \centering
      \setlength{\tabcolsep}{16pt}
    \scalebox{0.8}[0.8]{
    \begin{tabular}{lcc}
        \toprule
        \rowcolor{Gray} 
        Model & F1 & mAP \\
        \midrule
        CLIP \cite{Clip_radford} & 2.83 & 8.76\\
        Long-CLIP \cite{li2022longclip} & 10.01 & 3.17 \\
        MiniGPT \cite{minigptv2} & 18.45 & 16.47 \\
        LLaVA 1.5 \cite{Llava15} & 7.44 & 6.70 \\
         Florence \cite{florenceV2} & 16.04 & 20.99 \\
         \midrule
         \rowcolor{LightCyan}
        STING-BEE (Ours) & \textbf{34.69} & \textbf{29.80}\\
        \bottomrule
    \end{tabular}%
    }
    \label{tab_zeroshot_Scenecomprehension}
    \vspace{-1.5em}
\end{table}

\begin{table*}[t!]
    \centering
    \caption{\textbf{Cross-domain} performance comparison of different models on visual grounding tasks and referring threat localization.}
    \vspace{-1em}
    \setlength{\tabcolsep}{14pt}
    \scalebox{0.75}[0.75]{
    \begin{tabular}{lcccccccc}
        \toprule
        \rowcolor{Gray}
        Model & \multicolumn{2}{c}{Single-object grounding} & \multicolumn{2}{c}{Multi-object grounding} & \multicolumn{2}{c}{Refering Threat Localization} & \multicolumn{2}{c}{Overall} \\
        \cmidrule(lr){2-3} \cmidrule(lr){4-5} \cmidrule(lr){6-7} \cmidrule(lr){8-9}
& acc@0.5 & acc@0.25 & acc@0.5 & acc@0.25 & acc@0.5 & acc@0.25 & acc@0.5 & acc@0.25 \\
        \midrule
        MiniGPT-v2 \cite{minigptv2} & 6.5 & 13.61 & 2.7 & 7.02 & 5.7 & 12.54 & 4.9 & 11.05 \\
        Florence-2 \cite{florenceV2} & 7.8 & 16.74 & 3.2 & 7.84 & 6.3 & 13.23 & 5.7 & 12.60 \\
        \midrule
        \rowcolor{LightCyan}
        STING-BEE (Ours) & \textbf{12.5} & \textbf{28.75} & \textbf{4.8} & \textbf{11.76} & \textbf{8.9} & \textbf{24.03} & \textbf{8.7} & \textbf{21.51} \\
        \bottomrule
    \end{tabular}}
    \label{tab_detection_zeroshot}
    \vspace{-0.5em}
\end{table*}
 
\begin{figure*}[!t]
    \centering \small
    \includegraphics[width=\linewidth]{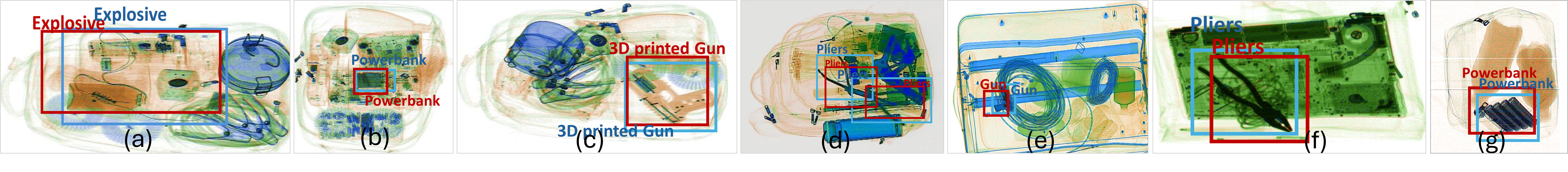} 
    \vspace{-2em}
    \caption{Qualitative examples from different scanners, including STCray (a-c), SIXray (d,e), and PIDray (f,g). STING-BEE effectively localizes a range of threats, such as \textit{3D-printed gun}, and demonstrates robust performance under challenging conditions, including occlusions (e.g., thick cables in (e) and electronic circuits in (b)) and significant intra-category variations (e.g., \textit{Powerbanks} in (b and g). \textcolor{blue}{Blue} boxes show the original bounding labels, while \textcolor{red}{red} boxes indicate model predictions.}
    \label{fig_Good_qual}
    \vspace{-1em}
\end{figure*}
\begin{figure}[!t]
    \centering \small
    \includegraphics[width=\linewidth]{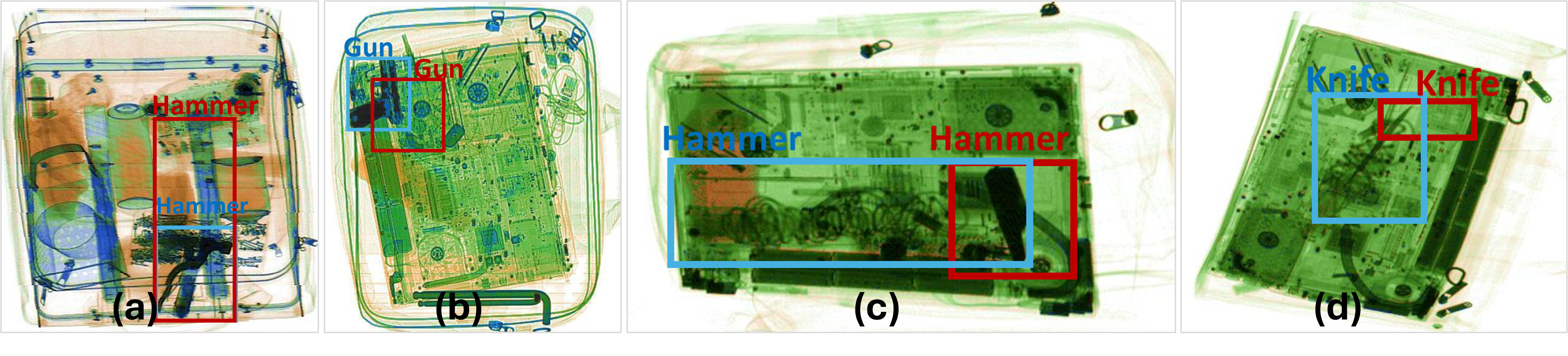} 
    \vspace{-2em}
    \caption{
    \textbf{Challenging examples:} STING-BEE recognizes heavily obscured objects e.g., \textit{Knife} in (d) hidden with cables and a laptop; however, its localization may be imprecise, detecting only part of the threat (e.g., knife tip in (d) or \textit{Hammer}-head in (c)).} 
    \label{fig_Bad_Qual}
\end{figure}
\begin{figure}[!t]
    \centering
    \includegraphics[width=\linewidth]{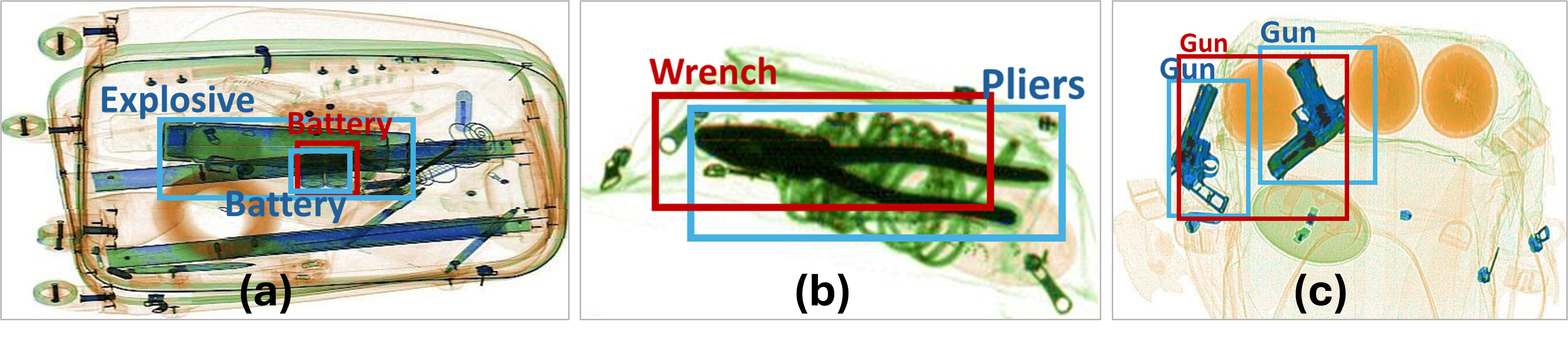} 
    \vspace{-2em}
    \caption{We showcase limitations of STING-BEE’s with multiple threats; (a) showing similar categories, (b) confusion between \textit{Wrench} and \textit{Pliers} and in (c) STING-BEE grouped threat items. \emph{Nonetheless, detecting any flagged threat prompts further inspection in real-world security applications.}}
    \label{fig_Ugly_Qual}
    \vspace{-1em}
\end{figure}
\subsection{Visual Question Answering}
\textbf{Evaluation Datasets:} In the absence of detailed captions in the existing X-ray dataset and established evaluation benchmarks for VQA in X-ray threat monitoring, we developed a comprehensive Visual Question Answers (VQAs) with 39,194 questions, including 22,920 questions from our STCray test set, 4,722 from SIXray, and 11,553 from PIDray. These include seven distinct question categories- Instance Identity, Instance Counting, Instance Location, Instance Attribute, Instance Interaction, Complex Visual Reasoning, and Misleading Questions, addressing critical aspects of visual comprehension relevant to security screening, spanning different levels of interleaved text and image understanding (see \autoref{supplement: Details on Visual Question Answers}). 
\textbf{Results:} We present VQA performance across different models in \autoref{tab_vqa_results}. For consistency, each model was evaluated using a multiple-choice format (A, B, C, D) (\autoref{supplement: Details on Visual Question Answers}). Our proposed model achieved the highest overall accuracy of 52.81\%, surpassing Florence-2 \cite{florenceV2}, MiniGPT \cite{minigptv2}, and LLaVA 1.5 \cite{Llava15}, particularly in categories requiring detailed reasoning and specific instance identification (see \autoref{supplement: Details on Visual Question Answers}). 

\subsection{Visual Grounding and Localization}
\textbf{Evaluation Datasets:} To assess visual grounding and referring threat localization capabilities, we compiled a comprehensive test set with 57,166 questions designed for referral tasks and 48,996 questions for grounding, using the detailed template (\autoref{tab_InstructioData_format}), from scans of SIXray \cite{sixray} and PIDray \cite{pidray}, (see \autoref{supplement: Details on Referring localization and Visual Grounding} for details). 
\textbf{Results:}  STING BEE achieves the highest, with 28.75\% for single threats, 11.76\% for multiple threats, and 24.03\% for referring threat localization, yielding an overall accuracy of 21.51\% (\autoref{tab_detection_zeroshot}). Notably, our model outperforms MiniGPT-v2~\cite{minigptv2} and Florence-2~\cite{florenceV2}. 
Despite limited training data, STING-BEE demonstrates strong grounding capabilities, underscoring the effectiveness of STCray dataset.
\subsubsection{STING-BEE: Strengths and Limitations}
STING-BEE demonstrates strong generalization across diverse datasets from multiple scanners, effectively detecting threats, including \textit{3D-printed guns} with faint outlines. It performs well under occlusion and with intra-category variations (\autoref{fig_Good_qual}). However, under severe occlusion (\autoref{fig_Bad_Qual}), the model recognizes objects, but its localization can be imprecise, covering only part of threat. This explains lower accuracy in \autoref{tab_detection_zeroshot}. Similarly, STING-BEE sometimes struggles with multiple threats e.g., in \autoref{fig_Ugly_Qual} where it detects \textit{Battery} but misses \textit{Explosive}, or it groups threat instances together or confusing similar categories (e.g., \textit{Wrench} vs. \textit{Pliers}). \emph{In practical X-ray threat detection,  identifying any threat suffices to flag the baggage}. Despite these challenges, STING-BEE achieves promising results e.g., state-of-the-art generalization against scanner-induced variability and intra-class differences, outperforming domain-specific models, including SDANet \cite{wang2021towards} and PSN \cite{eds_dataset} (designed for cross-domain threat localization) as shown in \autoref{tab_cross_domain_performance}. 
We hope it inspires further research in the community.

\begin{table}[t!]
    \centering \small
    \caption{We train all models on our STCray train set and evaluated them on a combined SIXray \cite{sixray} and PIDray \cite{pidray} dataset using only the common threat categories between these datasets. This allows us to evaluate vision-only models and our STING-BEE for scanner-induced variability and intra-class differences. Our model performs significantly better than vision-only models highlighting its cross-domain generalization.}
    \vspace{-1em}
    \centering
    \setlength{\tabcolsep}{8pt}
    \scalebox{0.8}[0.8]{
    \begin{tabular}{lcc}
        \toprule
        \rowcolor{Gray}
        Model & Accuracy@0.5 (\%) & Accuracy@0.25 (\%) \\
        \midrule
        Faster RCNN \cite{ren2017faster} & 1.5 & 4.9 \\
        DETR \cite{carion2020detr} & 1.3 & 4.5 \\
        \midrule
        PSN (CVPR 2022) \cite{eds_dataset} & 3.2  & 6.7 \\
        SDANet (ICCV 2021) \cite{wang2021towards} & 4.4 & 8.6 \\
        \midrule
        \rowcolor{LightCyan}
        STING-BEE (Ours) & \textbf{8.7} & \textbf{21.5} \\
        \bottomrule
    \end{tabular}%
    }
    \label{tab_cross_domain_performance}
    \vspace{-1.5em}
\end{table}

\section{Conclusion}
In response to the increasing complexity of aviation security, we introduce STCray, a multi-modal vision-language dataset capturing real-world threat concealment scenarios. STCray serves as a rigorous benchmark, revealing limitations in conventional and specialized X-ray models. Using STCray, we developed STING-BEE, the first domain-aware visual AI assistant for X-ray security, integrating scene comprehension, threat localization, visual grounding, and visual question answering, demonstrating strong potential for real-world aviation security applications.
\section{Acknowledgement}
This research was funded by Khalifa University under Grant Ref: CIRA-2021-052 and the ADNOC Research Fund (Ref: 21110553). The authors express their gratitude to Belal Irshaid, Manager of Lab Facilities, for his generous support. Syed Talal Wasim and Juergen Gall have been supported by the Federal Ministry of Education and Research (BMBF) under grant no. 01IS22094A WEST-AI and the ERC Consolidator Grant FORHUE (101044724).

{\small
\bibliographystyle{ieee_fullname}
\bibliography{Ref}
}

\clearpage
\appendix
\clearpage
\setcounter{page}{1}
\maketitlesupplementary

\begin{itemize}
    \item Additional details on STING Protocol (\autoref{supplement: Additional details on STING Protocol})
    \item Additional details on Instruction Tuning dataset (\autoref{supplement: Instruction following Dataset})
    \item Additional details on VQA Evaluation (\autoref{supplement: Details on Visual Question Answers})
    \item Additional details on Referring threat localization and Visual Grounding (\autoref{supplement: Details on Referring localization and Visual Grounding})
    \item Additional details on Data Augmentation (\autoref{supplement: CT-2-X-ray Augmentations})
    \item  STCray Data Characteristics (\autoref{supplement: STCray_Characteristics})
    \item STING BEE: Additional Results (\autoref{supplement:additional_results})
\end{itemize}

\section{Additional details on STING Protocol} 
\label{supplement: Additional details on STING Protocol}
\begin{figure}[t]
    \centering \small
\includegraphics[width=\linewidth]{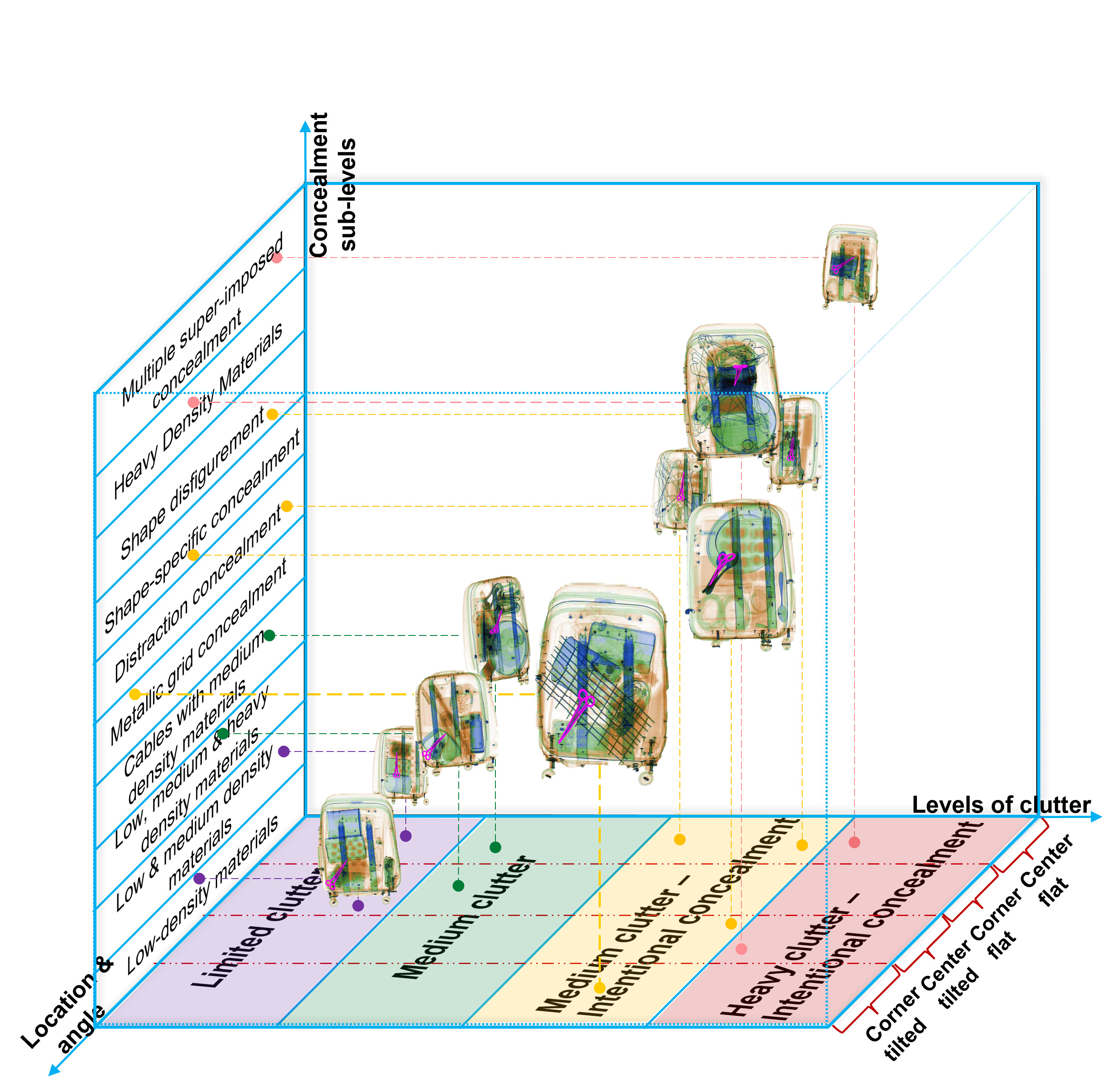} 
    \vspace{-2em}
    \caption{3D representation of STING protocol displaying the interplay between clutter levels, concealment sublevels, and location (from central to corner and flat to inclined) using \textit{Scissors} (with one sample for each sublevel for clarity).}
    \label{fig_DataColl}
    \vspace{-1em}
\end{figure}

The STING protocol underpins the STCray dataset, categorizing clutter into four levels—Limited, Medium, Heavy, and Extreme— progressively increasing occlusions and distractions. Concealment sublevels further diversify scenarios, ranging from low-density (e.g., organic items like books) to extreme configurations such as metallic grids and multi-layered superimposed materials. These concealments are further diversified by systematically varying the position and orientation of the threat, distorting threat appearances, and challenging model detection capabilities. 
\autoref{fig_DataColl}
 illustrates this interplay, showcasing the \textit{Scissors} across clutter, concealment, and positional variations.




\begin{algorithm}[ht]
\SetAlgoLined
\DontPrintSemicolon
\caption{Caption Generation using STING.}
\textbf{Input:} Threat items $T$, Concealment levels $L_c$, Locations $L_l$, Synonym sets $S_s$, X-ray scans $D_x$\\
\textbf{Output:} Generated captions $C_g$

\BlankLine
Initialize synonym pools $S_s$ for X-ray descriptors, verbs, positions, and concealment phrases;\\
Parse threat-specific metadata and mapping rules from $D_x$;\\
\BlankLine

\ForEach{image $I \in D_x$}{
    Extract:
    \begin{itemize}[noitemsep]
        \item Threat item $T(I)$;
        \item Concealment level $L_c(I)$;
        \item Location $L_l(I)$ (center or corner);
        \item Orientation $\Phi(I)$ (e.g., horizontal, tilted);
        \item Concealment phrase based on $L_c(I)$.
    \end{itemize}
    
    Select a random synonym for:
    \begin{itemize}[noitemsep]
        \item X-ray descriptor from $S_s$;
        \item Positioning verb from $S_s$;
        \item Location phrase from $S_s$.
    \end{itemize}
    
    \BlankLine
    \textbf{Construct Caption:}\\
    \Indp Combine elements into the caption:
    \vspace{-1em}
    \begin{center}
    \texttt{\small"\{X-ray descriptor\} \{positioning phrase\} a \{threat\}, \{concealment details\}, \{position\}."};
    \end{center}
    \Indm
    \vspace{-1em}
    Append constructed caption to $C_g$;
}
\BlankLine
\Return $C_g$;
\label{algo_caption_generation}
\end{algorithm}

\textbf{Caption Generation Using the STING Protocol}

The caption generation process, outlined in \autoref{algo_caption_generation}, leverages prior knowledge of threat-specific metadata and synonym sets to dynamically construct captions for each X-ray image \( I \in D_x \) collected using the STING Protocol. Synonym pools \( S_s = \{S_x, S_p, S_t, S_c\} \) provide linguistic variations for X-ray descriptors (\( S_x \)), positioning phrases (\( S_p \)), threat descriptions (\( S_t \)), and concealment details (\( S_c \)). For each image \( I \), attributes such as threat type \( T(I) \), concealment level \( L_c(I) \), location \( L_l(I) \), and orientation \( \Phi(I) \) are extracted to generate diverse and descriptive captions \( C_g \), enriching the STCray dataset for vision-language tasks.

Please note that our STING protocol utilizes predetermined metadata (e.g., threat type, placement, orientation) based on input from airport security personnel. Then it generates detailed captions directly from the metadata recorded during baggage preparation and scanning, eliminating human error. To validate the annotations, two independent annotators review a subset using the metadata and X-ray scans. We obtained a ROUGE-L value of 0.7 for captions, validating the annotations.
\begin{figure}[!t]
    \centering
    \includegraphics[width=0.85\linewidth]{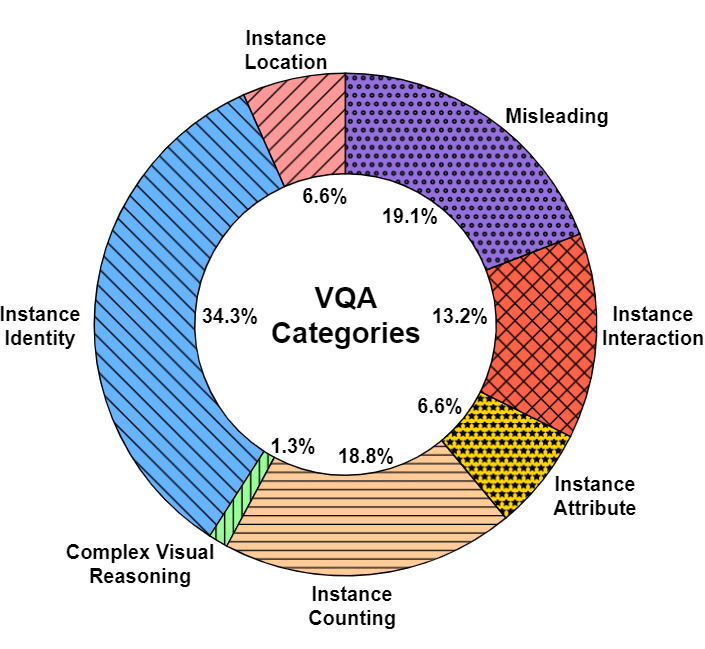} 
    \vspace{-1em}
    \caption{\textbf{VQA Evaluation Benchmark:} Distribution across seven categories to assess model robustness across diverse visual reasoning dimensions.}
    \label{fig_Plot_VQA_Categories}
    \vspace{-1em}
\end{figure}

\section{Additional details on Instruction Tuning dataset} \label{supplement: Instruction following Dataset}
\begin{figure*}[ht!]
\begin{tcolorbox}[colback=gray!10, colframe=gray!50, coltitle=black, width=\textwidth, boxrule=0.5mm, arc=4mm, left=2mm, right=2mm, top=2mm, bottom=2mm, boxsep=2mm]
\textbf{\large Prompt for VQA Instruction Generation:}  

You are an AI assistant analyzing X-ray baggage scans to detect prohibited items and security threats. Based on a description of the scan, answer questions as if you are visually analyzing the image. The description includes objects present in the scan, potential threat items, and objects placed to conceal them. Metallic items, such as guns, knives, and pliers, appear blue; organic items, such as 3D-printed guns and improvised explosives, appear orange; and inorganic items, such as circuits, powerbank, and battery, appear green. Using the description of the scan, design a conversation between you and a person asking about this scan, focusing on identifying threat items concealed within normal items.\\

The following are the threat categories likely to be present in the image alongside normal items: explosive, gun, 3D-printed gun, knife, bullet, syringe, battery, wrench, other sharp items, powerbank, scissors, hammer, pliers, and screwdriver. If none of the threat items are present, and only normal items are detected, the image is classified as "Nonthreat." Note that explosives can be intact or dispersed (dismantled). If dispersed, the description will mention the positions or concealment of the three main parts of the explosive: the container with explosive material, the circuit, and the battery. Sometimes the circuit, container, or battery may be expertly concealed within normal items.\\

Additionally, note that 3D-printed guns are difficult to detect because of their faint outlines, polymer-based structure, and orange appearance in the scan. You can include misleading questions about threat items that are not present and answer confidently that they are not present. Furthermore, tangled wires, cables, chains, stacked metallic items, circuits, and laptops may appear suspicious in the description. You can incorporate questions to clarify if there are any suspicious items in the image. Provide confident and definite answers, avoiding any uncertain or speculative responses.
\end{tcolorbox}
\caption{Prompt used for generating VQA instructions in STING-BEE. The prompt guides GPT-4 to generate conversations focusing on identifying concealed threat items in X-ray baggage scans, as if you are visually analyzing the scan.}
\label{fig_box_VQA_prompt}
\end{figure*}
\begin{figure}[!t]
    \centering
    \includegraphics[width=\linewidth]{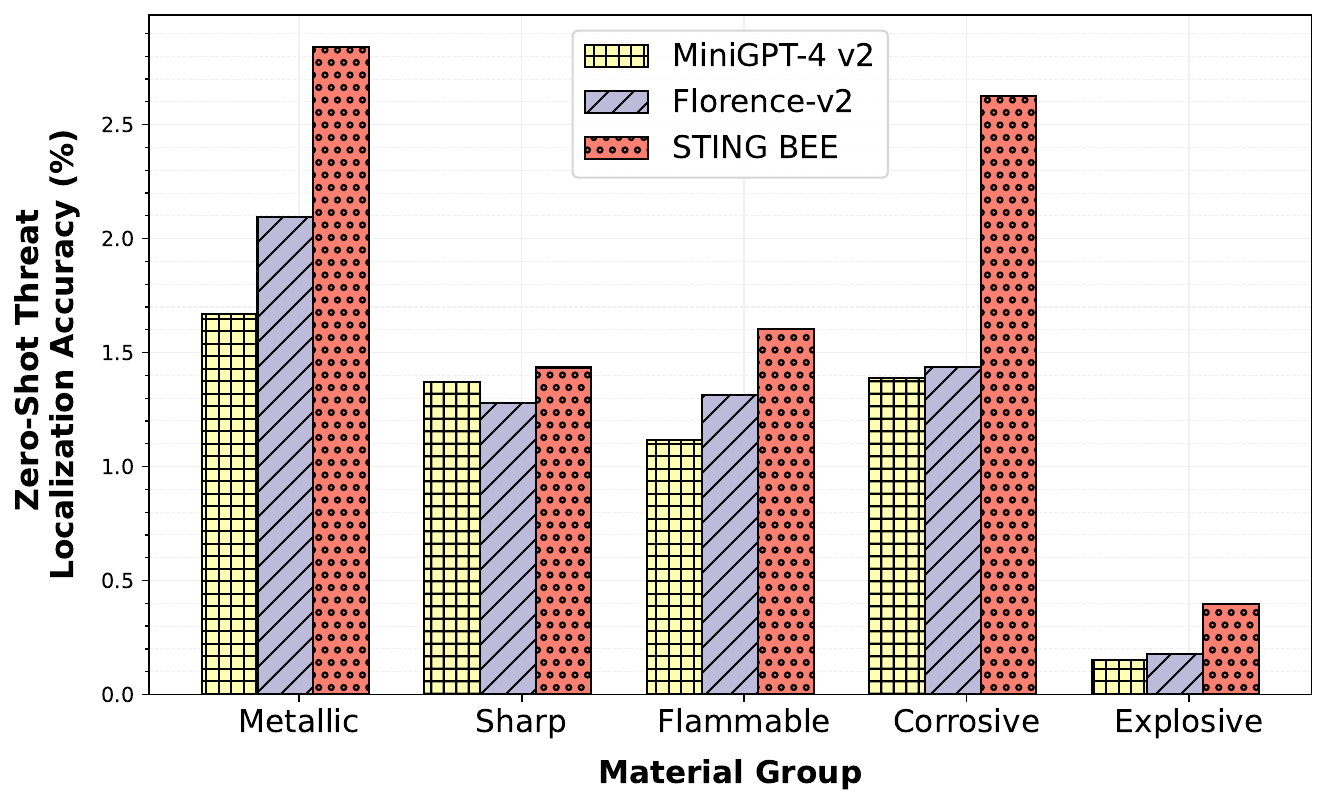} 
    \caption{Comparison of zero-shot threat localization accuracy of STING BEE, Florence-v2, and MiniGPT-4 v2 across material groups in localizing objects with diverse material characteristics.}
    \label{fig_Plot_Refer_Material}
\end{figure}

\begin{table}[!t]
\centering
\caption{\textbf{Question categories and sample questions:} Overview of the seven question categories used in the Visual Question Answering (VQA) dataset, comprising 39,194 questions across STCray, SIXray, and PIDray.}

\resizebox{\linewidth}{!}{
\begin{tabular}{>{\raggedright\arraybackslash}m{2cm}>{\raggedright\arraybackslash}m{8cm}}
\hline
\textbf{Evaluation Dimensions} & \textbf{Sample Question-Answer Pairs} \\ 
\hline
1. Instance Identity & What type of threat item is present in this X-ray image? \\
& A. Pliers B. Injection C. Lighter D. Battery \\
\hline

2. Instance Location & Where is the Battery located in this baggage X-ray scan? \\& A. Corner B. Middle C. Not Present \\
\hline

3. Instance Interaction & How is the Battery concealed in this baggage X-ray scan?  \\& A. Fully Covered B. Half Covered C. Uncovered \\
\hline

4. Instance Attribute & In what orientation is the Battery positioned within the baggage X-ray scan?  \\& A.Vertically B. Horizontally C. At an inclined angle. \\
\hline

5. Instance Counting & How many potential threats are present in this X-ray image? \\& A. One B. Two  C. Three D. None\\
\hline

6. Complex Visual Reasoning &How can you infer that there is a metal gun in the baggage scan?  \\& A. Since the gun is bluish or greenish, it can be easily inferred as metal gun. B. Since the gun is orange-coloured, it can be easily inferred as metal gun. C. None of the above. \\
\hline

7. Misleading & Where is the Cutter located in the baggage scan? \\ & A. There is no cutter in the image. B. Toward the corner of the image. C. In the middle of the image.\\
\hline
\end{tabular}}

\label{tab_vqa_sample_questions}
\end{table}

\begin{figure*}[ht!]
    \centering
    \begin{tcolorbox}[colback=gray!10, colframe=gray!50, coltitle=black, width=\textwidth, boxrule=0.5mm, arc=4mm, left=2mm, right=2mm, top=2mm, bottom=2mm, boxsep=2mm]
    
    \includegraphics[width=0.3\textwidth]{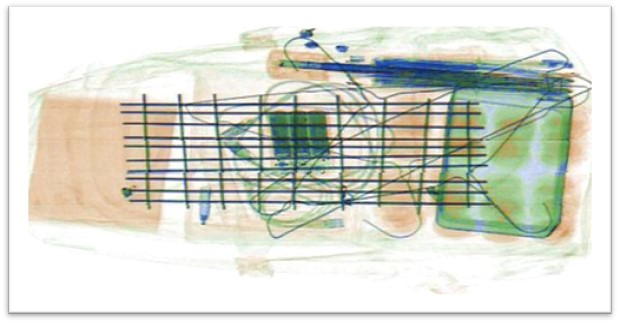} \\
    \textbf{STCray Caption:} \\
    \textit{"X-ray scan showing a power bank aligned horizontally, in the middle of a travel bag, with the power bank covered by the metal grid and hangers, along with some random cables for distraction, and there are other items like an umbrella and other normal passenger items."} \\
    
    \textbf{Multi-Round Question-Answer Conversation:}
    \begin{itemize}[noitemsep, left=0pt]
        \item \textbf{Human:} Can you identify any potential security threats in this X-ray scan?
        \item \textbf{Assistant:} Yes, the scan contains a powerbank, which is a prohibited item.\\

        \item \textbf{Human:} Where is the powerbank located in the scan?
        \item \textbf{Assistant:} The powerbank is aligned horizontally in the middle of the travel bag.\\

        \item \textbf{Human:} What is obscuring the powerbank in the scan?
        \item \textbf{Assistant:} The powerbank is covered by a metal grid, hangers and various cables, which serve as distractions.\\

        \item \textbf{Human:} What color and shape should I look for to identify the powerbank?
        \item \textbf{Assistant:} You should look for a greenish rectangular block, as powerbanks appear with a greenish tinge due to their inorganic material.\\

        \item \textbf{Human:} Are there any other items in the scan that could be mistaken for a threat?
        \item \textbf{Assistant:} Yes, there are other normal items such as an umbrella, hangers and random cables, which could distract from identifying the powerbank's presence.\\

        \item \textbf{Human:} How do these normal items affect the detection of the powerbank?
        \item \textbf{Assistant:} The presence of normal items like cables and an umbrella can create confusion and distract from the identification of the powerbank.\\

        \item \textbf{Human:} What should be done about the powerbank in this scan?
        \item \textbf{Assistant:} The powerbank should be flagged for further inspection and potential removal from the baggage, as it is a security threat.
    \end{itemize}
    \end{tcolorbox}
    \caption{An instance of GPT-4 generated VQA instruction featuring the baggage scan and its corresponding caption from STCray at the top. The instruction-following data, generated using only the caption, is displayed below. Note that the baggage scan was not provided to GPT-4 during prompting and is included here solely for readability.}
    \label{fig_vqa_sample}
\end{figure*}

The Visual Question-Answering (VQA) instructions were generated using GPT-4 \cite{gpt4} for training STING BEE. These conversations, derived from STCray training set captions, focused on threat identification, complex reasoning, and positional queries. A carefully designed GPT prompt (\autoref{fig_box_VQA_prompt}) guided the generation process, ensuring task-specific queries and precise responses. \autoref{fig_vqa_sample} illustrates an example of a multi-round VQA conversation paired with its associated baggage scan and caption.

\section{Additional details on VQA Evaluation} 
\label{supplement: Details on Visual Question Answers}
To evaluate STING-BEE's visual reasoning capabilities, we developed a comprehensive VQA dataset, drawing inspiration from the SEED Bench \cite{seedBench} methodology. It comprises 39,194 questions derived from STCray, SIXray, and PIDray. These questions span seven distinct categories, as shown in \autoref{fig_Plot_VQA_Categories}, targeting critical dimensions of visual reasoning.

\begin{itemize}[noitemsep, topsep=0pt, left=0pt]
    \item \textbf{Instance Identity}: Tests the model's ability to recognize and classify specific threat types (e.g., gun vs. knife), requiring image-wide context.
    \item \textbf{Instance Counting}: Evaluates the model's capacity to count potential threats in a scene, crucial for understanding scenario complexity.
    \item \textbf{Instance Location}: Assesses spatial understanding by determining object locations (e.g., center vs. corner).
    \item \textbf{Instance Attribute}: Focuses on identifying object-specific features like orientation or occlusion.
    \item \textbf{Instance Interaction}: Requires object-level reasoning and examines how objects interact, such as occlusion.
    \item \textbf{Complex Visual Reasoning}: Requires the model to infer threats from contextual cues, emphasizing domain-specific higher-order reasoning.
    \item \textbf{Misleading Questions}: Includes deliberately misleading queries to evaluate model precision.
\end{itemize}
Sample questions from each category are presented in \autoref{tab_vqa_sample_questions}.

\begin{figure*}[!t]
    \centering \small
    \includegraphics[width=\linewidth]{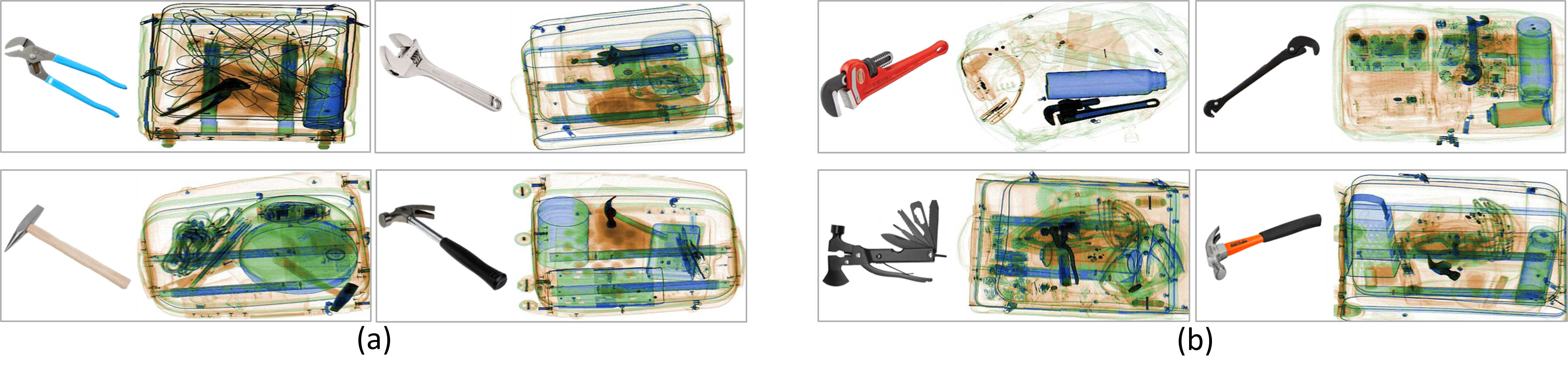} 
    \caption{Illustration of train-test diversity and intra-category variability in the STCray dataset. The top row shows examples from the \textit{Wrench} category: (a) \textit{Pliers Wrench} and \textit{Crescent Wrench} from the train set, and (b) \textit{Pipe Wrench} and \textit{Self-adjusting Wrench} from the test set. The bottom row highlights the \textit{Hammer} category: (a) \textit{Brick Hammer} and \textit{Claw Hammer} from the train set, and (b) \textit{Multi-tool Hammer} and \textit{Framing Hammer} from the test set. A clear separation between train and test sets underscores the high intra-category variance, simulating real-world diversity, and challenging model generalization.}
    \label{fig_Train_test_diversity}
    \vspace{-1em}
\end{figure*}
\begin{figure*}[htbp]
    \centering \small
    \includegraphics[width=0.8\linewidth]{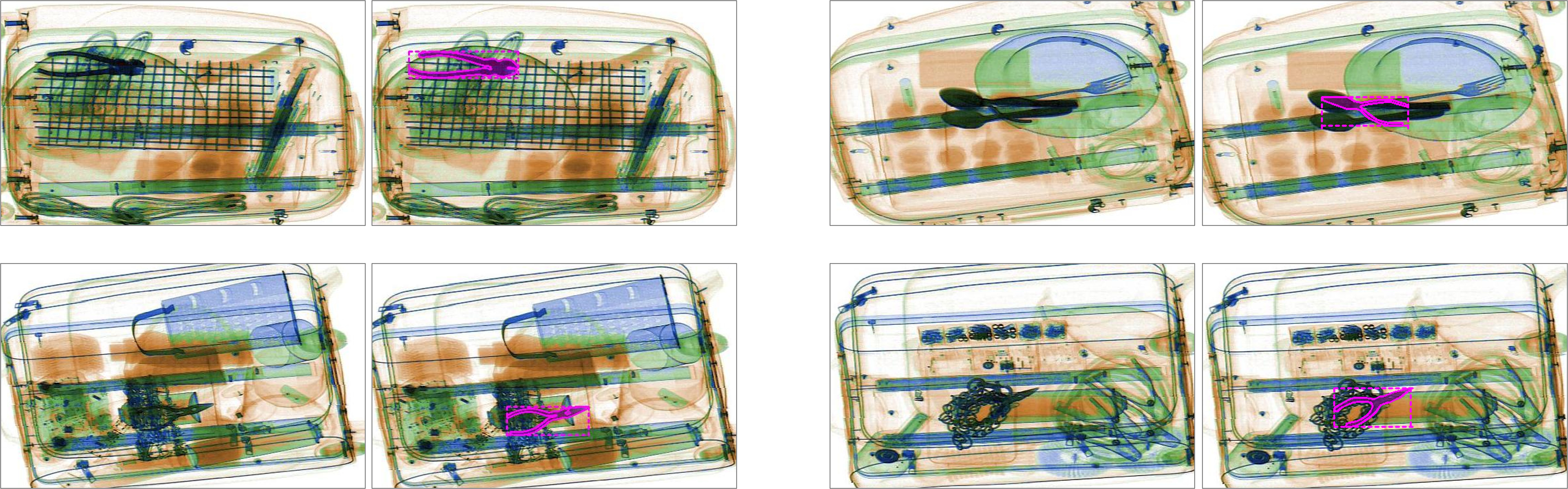} 
    \caption{Illustration of diverse occlusion strategies in the STCray dataset across the train and test subsets. The training set (top row) employs techniques such as occluding the \textit{Plier} with metallic grids (left) and disguising it with spoons (right). The testing set (bottom row) introduces novel strategies, including obscuring the \textit{Plier} with a box of Integrated Chips (ICs) (left) and disguising it as an ornament using a chain (right).}
    \label{fig_Train_test_occlusion}
    \vspace{-1.5em}
\end{figure*}
\begin{figure*}[t]
    \centering
      \includegraphics[width=0.5\linewidth, height=3cm]{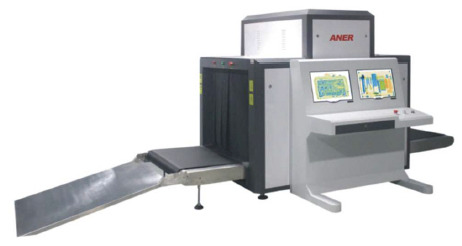} 
    \includegraphics[width=0.3\linewidth , height=3cm]{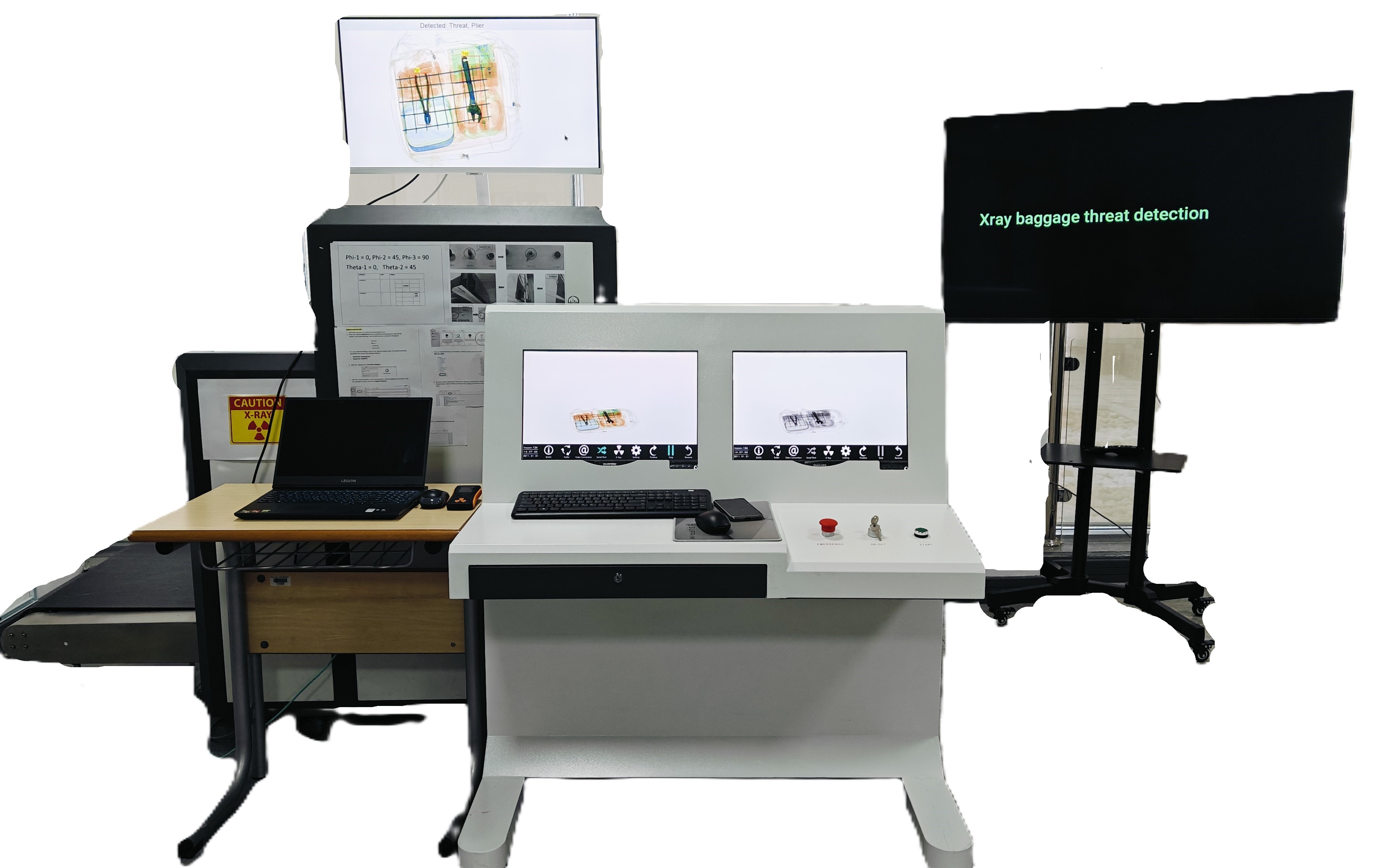} 
       \includegraphics[width=0.19\linewidth, height=2cm]{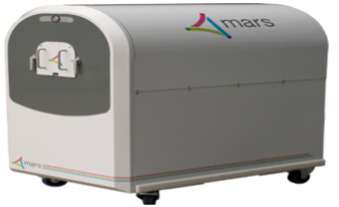} 
    \caption{The ANER K8065 X-ray Baggage Scanner used for the STCray dataset collection. On the right: the MARS Microlab 5X120 CT scanner which was employed for data augmentation.}
    \label{fig_scanner}
\end{figure*}
\begin{figure}[!t]
    \centering \small
    \includegraphics[width=\linewidth]{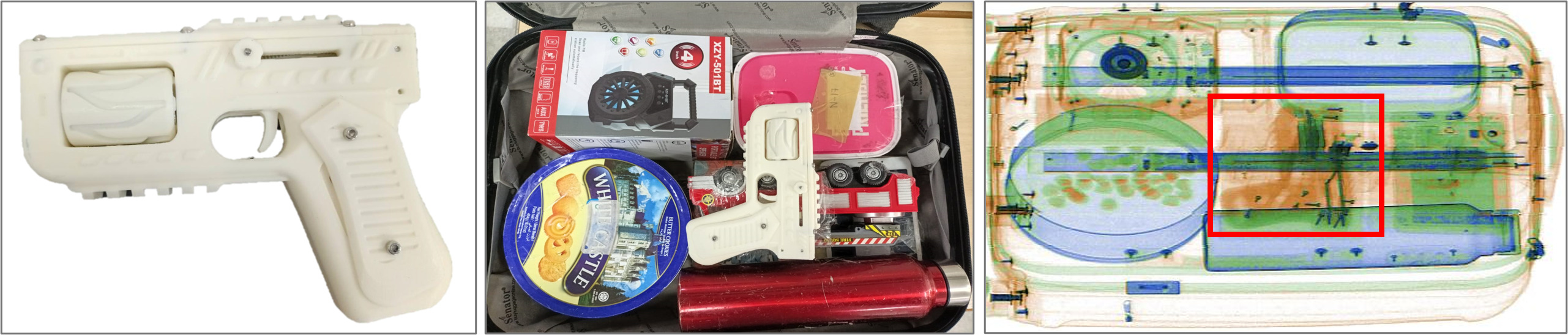} 
    \caption{
    Left:
    \textit{Maverick} 3D-printed firearm. Center: A sample baggage setup, concealing the firearm among everyday items. Right: The corresponding X-ray scan, illustrating the challenge of detecting faint outlines and identifying the firearm in a cluttered, realistic baggage scenario.}
    \label{fig_3D_gun}
    \vspace{-1em}
\end{figure}
\begin{figure}[!t]
    \centering \small
    \includegraphics[width=\linewidth]{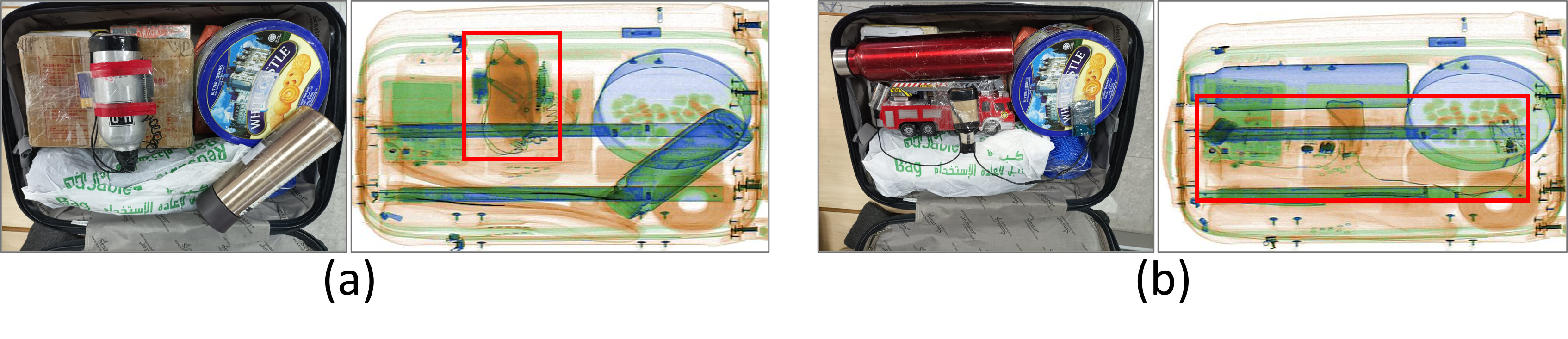} 
    \caption{Samples from the \textit{Explosive} category: baggage preparation and the corresponding X-ray scan. (a) Cohesive (compact) IED with components grouped together, and (b) IED with components distributed across the baggage and connected by wires.}
    \label{fig_Explosive_intact_dispersed}
\end{figure}
\section{Additional details on Referring threat localization and Visual Grounding}
\label{supplement: Details on Referring localization and Visual Grounding}
STING-BEE's performance in visual grounding and referring threat localization was evaluated using SIXray and PIDray datasets. In SIXray, which includes five threat categories, we grouped threats into two material groups: \textit{Metallic} (\textit{Gun}, \textit{pliers}, \textit{wrench}) and \textit{Sharp} (\textit{Knife}, \textit{Scissors}) to enable a more granular evaluation of model performance across different material types. While PIDray's categories were organized into five groups: \textit{Metallic}, \textit{Sharp}, \textit{Flammable}, \textit{Corrosive}, and \textit{Explosive}. Evaluation questions followed templates like \texttt{\small[grounding] Describe the baggage scan.} and \texttt{\small[refer] Please find <p>the threat category</p>}, with ground truth mapped to normalized bounding boxes for consistency. As shown in \autoref{fig_Plot_Refer_Material}, STING-BEE outperformed MiniGPT-v2~\cite{minigptv2} and Florence-2~\cite{florenceV2}, demonstrating robust performance across diverse material groups.

\section{Additional Details on Data Augmentation}\label{supplement: CT-2-X-ray Augmentations}

Existing approaches on data augmentation  \cite{MeryKatsaggelos,duan2023_DA} use single-direction projection methods limiting realism and orientation diversity. For the STCray dataset, our data augmentation approach is distinguished by a novel multi-view projection aspect enabled by our in-house CT scanner (see \autoref{sec:datacollection}). Using this scanner, we generated three CT scans of threat objects, rotating them in various orientations and projecting them into 2D X-ray images through a carefully designed technique. These X-ray images simulate the scanning of threat items from multiple perspectives. Subsequently, we integrated these scans into normal baggage scans using standard fusion and colourization techniques. The details of the approach are provided next.

\noindent Given $M(x,y,z)$ a 3D tensor representing a CT scan of a threat object, we apply a rotation $R_{\phi, \theta, \psi}(M)$, where $\phi, \theta, \psi$ refer to the  Euler angles representation. Afterwards, assuming an ideal narrow beam geometry and ignoring 
scatter, we approximate the projection of the threat item by computing the integration along the nominal z-axis\cite{hsieh2009}:
$P(x,y) = \int R_{\phi, \theta, \psi}(M(x,y,z))dz$.  The final image intensity $I$ at any point is then calculated using the exponential attenuation model\cite{hsieh2009}:
\begin{equation}
\label{equ:volproj}
I = I_0 e^{-\int R_{\phi, \theta, \psi}(M(x,y,z))dz}
\end{equation} 
where $I_0$ is the initial beam intensity being exponentially attenuated by the projection path.  

\noindent 
In the final stage, we fused a patch from a normal baggage scan with a threat image at a given location using pixel-wise multiplication. Both the baggage scan and the threat image are represented in grayscale. For colorization, we utilized a UNet architecture trained in a self-supervised manner to convert grayscale images into pseudo-colored scans. However, more advanced colorization techniques can be used, such as the recent method proposed by Duan et al. \cite{duan2023_DA}.

In summary, our method generates 2D projections with varying X-ray attenuation levels, introducing increased variability in the augmented scan data. This enhanced diversity effectively simulates real-world smuggling scenarios where threat items are positioned in unconventional ways within baggage to improve concealment. 
Furthermore, our approach enables automated annotation by implicitly generating threat masks within the augmented data.

\begin{figure*}[t!]
    \centering
    \includegraphics[width=0.8\linewidth]{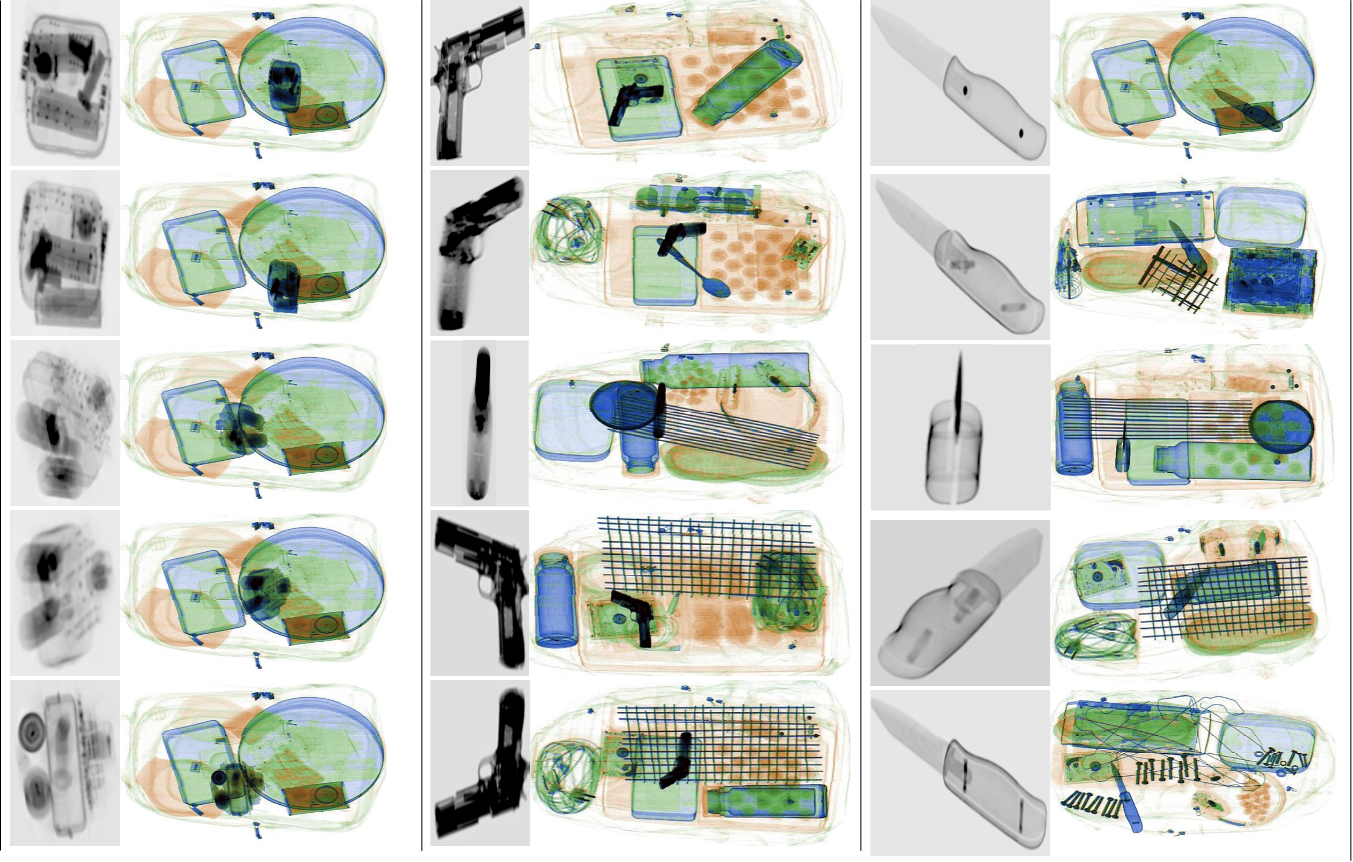}
     \includegraphics[width=0.8\linewidth]{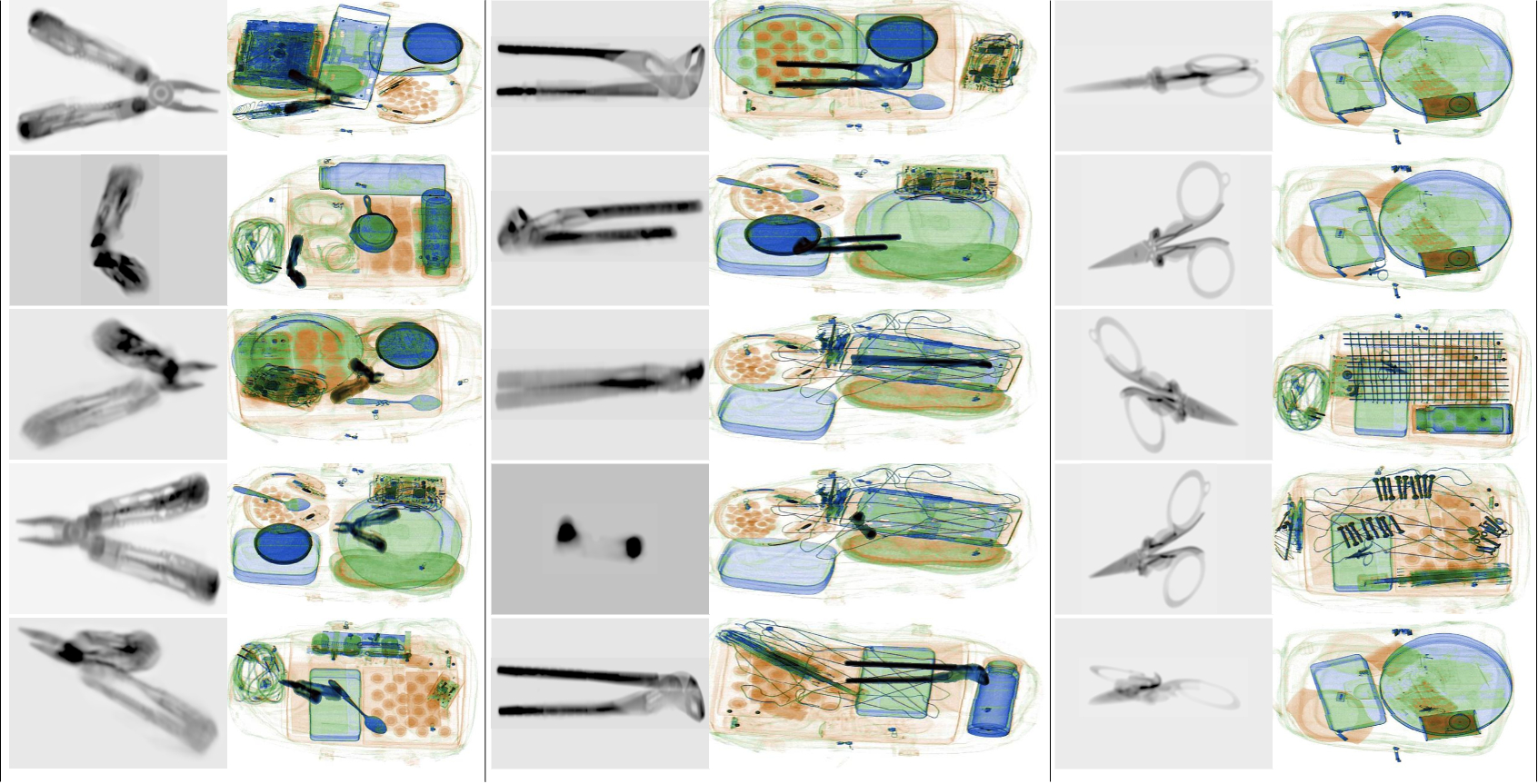}
    \caption{\small 2D X-ray projections, at different angles, from CT scans of threat items (explosives, guns, knives, pliers, wrenches, and scissors) and their augmented images. Notice the variation in the grayscale across the different projections reflecting different levels of X-ray attenuation. Note also that the atypical poses of threats generated through these projections create challenging instances, making threat detection more difficult. For example,  the fourth and last instances of the wrench (3rd column) and the scissors (last column), respectively.}
    \label{fig_augmentation}
\end{figure*}

\noindent Due to the limited size of our CT scanner's imaging chamber (as detailed in the equipment section \autoref{sec:datacollection}), we initially produced augmented scan data for the six threat items reported in \cref{fig_augmentation}
. 


\section{STCray Dataset Characteristics}
\label{supplement: STCray_Characteristics}
\begin{figure*}[t]
    \centering \small
\includegraphics[width=16cm, height=18cm]{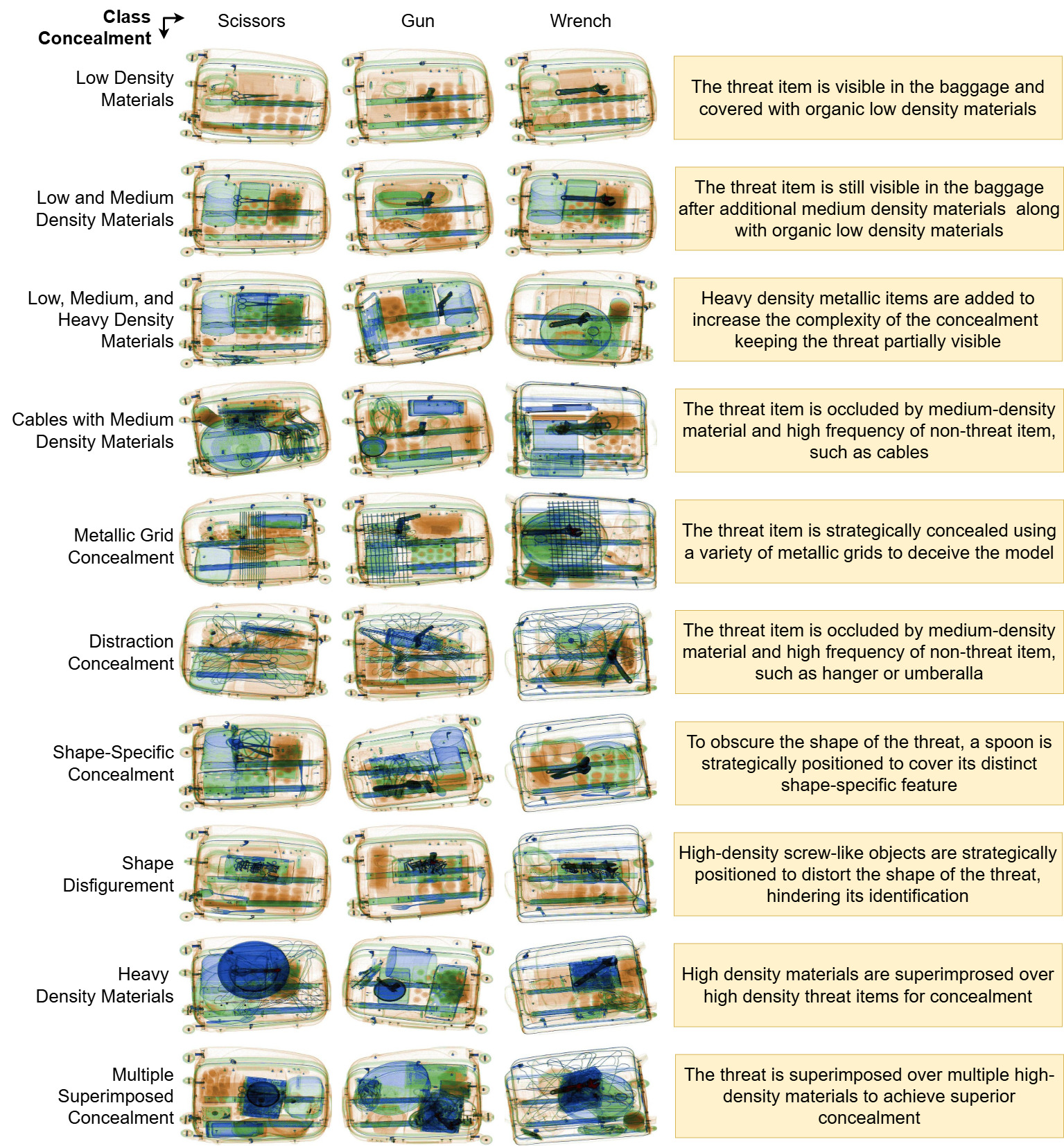}
    \caption{Illustrative examples of the 10 concealment levels utilized in the STCray dataset, demonstrating increasing complexity in obscuring threat items (scissors, gun, wrench) through diverse materials and strategies. The accompanying descriptions explain the concealment techniques and their effectiveness in reducing threat visibility.}
  \label{fig_concealment_levels}   
\end{figure*}

The \emph{Strategic Threat Concealment X-ray (STCray)} dataset introduces unique challenges that set it apart from existing X-ray security benchmarks, establishing itself as a pivotal resource for advancing research in baggage threat detection. Below, we provide a comprehensive overview of the characteristics and challenges that STCray brings to the research community.
\begin{itemize}[noitemsep, left=0pt]
\item \textbf{Emerging Sophisticated Threats:}STCray addresses modern security challenges by incorporating a diverse range of contemporary threats, including 3D-printed firearms and improvised explosive devices (IEDs), alongside other prohibited items. Specifically, the dataset features three distinct 3D-printed gun designs: the single-shot \textit{Liberator} \cite{liberator}, the minimalist and recent \textit{Harlot} \cite{harlot_url}, and the pepper-box styled \textit{Maverick} \cite{maveick}. These designs were chosen to represent diverse shapes and structures, reflecting the variety encountered in real-world scenarios. Detecting 3D-printed firearms is particularly challenging due to their faint outlines and unconventional material properties, which blend with benign objects in X-ray scans (see \autoref{fig_3D_gun}).

Similarly, IEDs pose unique detection challenges due to their non-standard shapes and multi-component configurations. As depicted in \autoref{fig_Explosive_intact_dispersed}, STCray incorporates both cohesive (compact) and dispersed IED types: cohesive designs consolidate explosive charges, detonators, and power sources into a single, unified threat, while dispersed designs distribute these components throughout the baggage, interconnecting them with wires. These variations further complicate detection, mirroring the diverse tactics employed in real-world scenarios.
\item \textbf{Strategic Threat Concealment: }STCray is meticulously designed to reflect the complexities encountered in real-world baggage screening scenarios, using a carefully designed STING protocol , incorporating systematic threat concealment strategies that mimic smuggling tactics and concealment practices. Our collaborators from baggage screening units at the airport shared these smuggling practices with us.
Threat items are deliberately positioned and occluded by clutter and dense objects, with varying levels of overlap, material density, and angular placement. This systematic approach ensures that the dataset captures realistic concealment scenarios, pushing models to their limits. \autoref{fig_concealment_levels} reports examples of different concealment levels for three different threat items. 
\item \textbf{Diversity and Realism: }Unlike traditional datasets that often retain similar configurations across train and test sets—varying primarily in occlusion levels but preserving the same benign objects and threat items—STCray was designed to reflect the complexity and variability of real-world baggage scenarios. This diversity ensures realistic and challenging evaluation scenarios, mimicking the complexity and variability of real-world baggage screening (see \autoref{supplement:Intra-category}).
\end{itemize}

\subsection{Intra-Category Variance} 
\label{supplement:Intra-category}
The train and test subsets of the STCray dataset both adhere to the STING protocol, ensuring systematic threat concealment and realistic variations.  However, to introduce real-world relevance and challenge model generalizability, we have ensured intra-category diversity and diverse concealment strategies across the train and test subsets. For example, as shown in \autoref{fig_Train_test_diversity}, the train set (top row) includes baggage scans featuring \textit{Wrench} instances such as the \textit{Pliers Wrench} and \textit{Crescent Wrench} (left), while the test set contains entirely different \textit{Wrench} types, such as the \textit{Pipe Wrench} and \textit{Self-adjusting Wrench} (right). Similarly, in the \textit{Hammer} category (bottom row), the train set includes the \textit{Brick Hammer} and \textit{Claw Hammer} (left), while the test set introduces novel instances like the \textit{Multi-tool Hammer} and \textit{Framing Hammer} (right).

This diversity extends to the occlusion strategies employed across subsets. For instance, as shown in \autoref{fig_Train_test_occlusion}, the training set (top row) uses tactics such as metallic grids to occlude the \textit{Plier} or spoons to disguise it. In contrast, the testing set ( bottom row) introduces entirely new strategies, including obscuring the \textit{Plier} with a box of Integrated Chips (ICs) or disguising it as an ornament using a chain.

These variations in both object structure and occlusion highlight STCray's emphasis on high intra-category variance and its ability to challenge models beyond the limitations of conventional datasets, where train and test splits often lack such diversity. By integrating these realistic complexities, STCray not only mimics real-world scenarios but also serves as a robust benchmark for evaluating model generalization and resilience to real-world variability, thereby addressing a critical gap in existing X-ray security datasets.

\subsection{Scanner Details} \label{sec:datacollection}
The STCray dataset is collected using an ANER K8065 X-ray scanner \autoref{fig_scanner}. It is an advanced X-ray baggage scanner designed for security applications in airports, metro stations, and other security checkpoints. It features a tunnel size of 800 mm in width and 650 mm in height, accommodating a wide range of luggage sizes. The conveyor operates at an adjustable speed of 0.22 meters per second and can handle a maximum load of 200 kilograms. It is equipped with a high-resolution 17-inch LCD display and utilizes a dual-energy L-type photoelectric diode array detector with 12-bit depth, providing precise and detailed images. The scanner is equipped with software providing image processing capabilities like edge enhancement, super image enhancement, high and low penetration display modes, and a magnifier function for partial enlargement. The machine boasts up to 40 mm steel penetration capability, effectively capturing concealed items within dense materials. 

\noindent For the data augmentation, we used our in-house CT scanner model MARS Microlab 5X120 (\cref{fig_scanner}), which offers a spatial resolution range of [50–200 $\mu$ mm]. Its chamber can accommodate objects with a maximum diameter of 100 mm and a length of 350 mm. Consequently, we were able to scan only a subset of the available threat items, including an explosive, a gun, a knife, pliers, a wrench, and scissors.
\section{STING BEE: Additional Results}
\label{supplement:additional_results}
\begin{figure}[!t]
    \centering    \includegraphics[width=\linewidth]{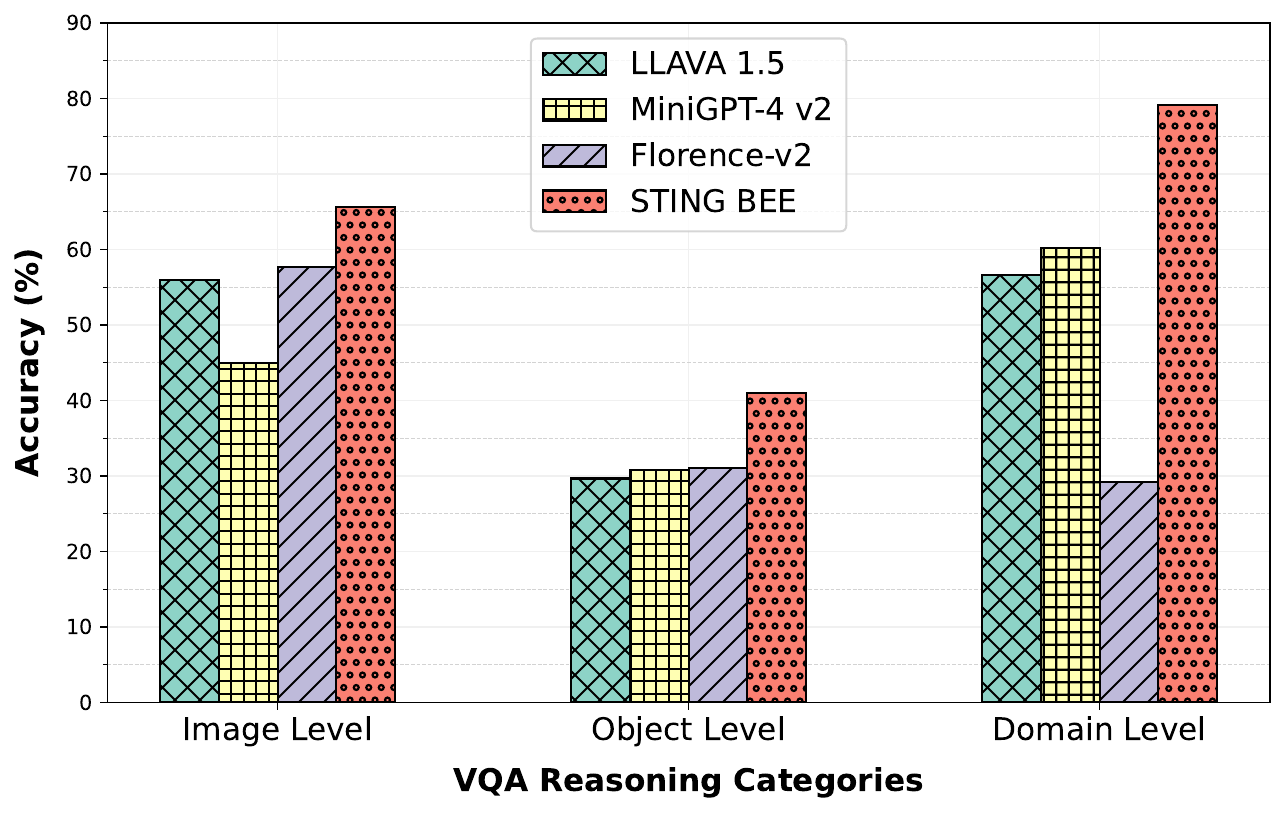} 
    \caption{Model Accuracy Across Different Reasoning Categories in Visual Question Answering for X-ray Threat Detection.}
    \label{fig_plot_vqa}
\end{figure}
\subsection{VQA Capabilities}
To assess reasoning capabilities, we grouped the VQA categories by reasoning levels—Image-Wide Context (Instance Identity, Instance Counting, Instance Location), Object-Level Reasoning (Instance Attribute, Instance Interaction), and High-Level Domain Awareness (Complex Visual Reasoning)—and plotted model accuracy across these groups (see \autoref{fig_plot_vqa}). Further analysis of the plot shows that STING BEE outperforms other models in both Image-Wide (65.64\%) and domain-level (79.17\%) tasks, demonstrating its comprehensive image-wide and domain-specific knowledge due to domain-targeted instruction tuning across multiple tasks.

\begin{table}[t!]
    \centering \small
    \caption{\textbf{Cross-Domain} comparison of different models on multi-class classification tasks. Models are trained on the STCray training set and tested on a combined SIXray and PIDray dataset, showing overall mAP and F1-score.}
    \vspace{-1em}
    \setlength{\tabcolsep}{15pt} 
    \scalebox{0.85}[0.85]{ 
    \begin{tabular}{lcc}
        \toprule
        \rowcolor{Gray}
        Model & mAP & F1 Score \\
        \midrule
        EfficientNet-B4 \cite{tan2019efficientnet} & 9.5 & 9.0 \\
        GWFS \cite{hassan2023unsupervised}         & 5.3 & 6.6 \\
        DSACDIC \cite{zhao2024deep}               & 3.7 & 6.7 \\
        ViT \cite{vit}                            & 8.4 & 12.1 \\
        DeiT \cite{touvron2021training}           & 9.1 & 8.9 \\
        DINO \cite{caron2021emerging}             & 11.4 & 8.3 \\
        CLIP \cite{Clip_radford}                  & 8.1 & 3.3 \\
        LongCLIP \cite{li2022longclip}            & 9.5 & 4.7 \\
        \rowcolor{LightCyan}
        \textbf{STING BEE (Ours)}                 & \textbf{30.1} & \textbf{36.2} \\
        \bottomrule
    \end{tabular}}
    \label{tab_Cross-Domain-Generalization}
    \vspace{-0.5em}
\end{table}

\subsection{Cross-Domain Generalization}
The results in \autoref{tab_Cross-Domain-Generalization} demonstrate the generalization ability of our proposed model, STING BEE, in multi-class classification tasks. To ensure a fair evaluation, all state-of-the-art and baseline models were trained on the STCray dataset and tested on a combined SIXray and PIDray dataset, with unique categories such as \textit{Baton} and \textit{Sprayer} removed to align the evaluation settings. Despite the competitive results from GWFS and DSACDIC, which are designed for cross-domain scenarios, on specific classes, STING BEE achieved the highest overall mAP of 30.1 and F1 score of 36.2, consistently outperforming all other models across most threat items, including Gun, Wrench, Hammer, and Powerbank. These results highlight STING BEE’s ability to handle scanner-induced variability and intra-class differences, establishing it as a robust benchmark for cross-domain generalization in threat detection.

\subsection{Comparison with General-purpose VLMs}
Since STING-BEE is the first X-ray security visual AI assistant capable of scene comprehension, visual grounding, and VQA, we compared it with other general-purpose (open and closed-source) VLMs. The results in \autoref{tab_Qwen_Llama_STINGlimited_GPT4} show that STING-BEE outperforms GPT-4o on both VQA and cross-domain scene comprehension tasks. STING-BEE also surpasses Qwen2-VL and Llama 3.2 in scene comprehension, VQA, and grounding capabilities, highlighting the need for domain-aware VLMs in X-ray security screening. To further analyze the relevance of the detailed captions in STCray, we trained STING-BEE with only image labels and box-level annotations. As shown in \autoref{tab_Qwen_Llama_STINGlimited_GPT4}, STING-BEE with limited annotations still performs better than general purpose models; however, performs lower compared to STING-BEE trained with detailed annotations (\autoref{tab_Qwen_Llama_STINGlimited_GPT4}).

\begin{table}[t]
\centering
\caption{\scriptsize Our STING-BEE outperforms Llama 3.2, Qwen2-VL,  LLaVa (Finetuned) and GPT-4o. IL:Instance Location, CR:Complex Reasoning, IID:Instance Identity, IC:Instance Counting, M:Misleading, IA:Instance Attribute, II:Instance Interaction.
}
\vspace{-1em}
\setlength{\tabcolsep}{1.9pt}
\scalebox{0.5}[0.5]{
\begin{tabular}{@{}lcccccccccccc@{}}
\toprule
\rowcolor{Gray}.
\textbf{Model} & \multicolumn{2}{c}{\textbf{Scene Comprehension}} & \multicolumn{8}{c}{\textbf{Visual Question Answering (VQA)}} & \multicolumn{2}{c}{\textbf{Grounding}} \\ 
\cmidrule(lr){2-3} \cmidrule(lr){4-11} \cmidrule(lr){12-13}
               & F1   & mAP  & IL & CR & IID & IC & M & IA & II & Overall & acc@0.5 & acc@0.25 \\ \midrule
GPT-4o& 19.4  & 18.1  &  21.5   &  16.2    &  36.3  &  29.6   &  25.2  & 19.02  &   18.3   &   31.3& - &-  \\     LLaVa 1.5 (Finetuned)&28.3& 22.2 & 35.2 & 59.4 & 77.8 &36.1& 15.9 & 48.2 &  25.3 & 45.1&-&-    \\         
Llama 3.2 & 13.4 & 17.3 &  26.5 &  29.3 &  36.3 &   20.8 &  11.6&   25.4 &  13.3 &  23.5 & 1.9 &5.3\\ 
Qwen2-VL&18.6&16.9&31.3&15.1&41.3&24.1&18.8&36.1&13.1&27.8&1.2&3.9\\
\begin{tabular}[l]{@{}l@{}}STING-BEE \\(limited captions)\end{tabular}&22.1&18.2&25.3&41.3&69.2&35.4&20.5&42.2&27.6&41.8&5.3&14.1\\
\midrule
STING-BEE&\textbf{34.7}& \textbf{29.8} & \textbf{49.2} & \textbf{79.2} & \textbf{80.0}  &  \textbf{45.2} & \textbf{27.8} & \textbf{52.8} & \textbf{35.0}  & \textbf{52.8}&\textbf{8.7}&\textbf{21.5} \\ 
\bottomrule
\end{tabular}
}
\label{tab_Qwen_Llama_STINGlimited_GPT4}
\vspace{-1em}
\end{table}

\subsection{Qualitative Results}
\label{supplement:qualitative}
\begin{figure*}[t!]
    \centering
    \includegraphics[width=1\linewidth]{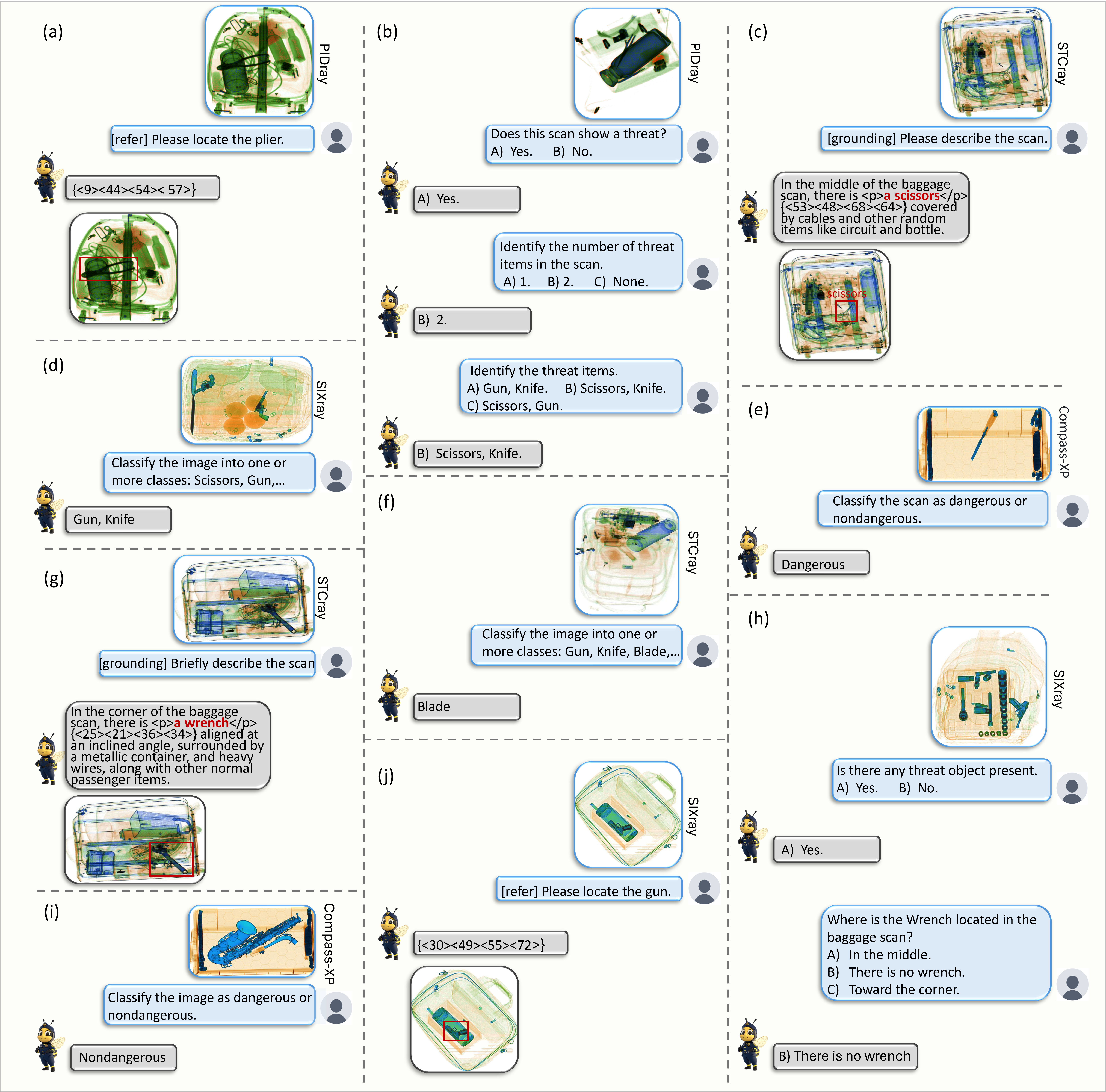} 
    \caption{Qualitative examples showcasing the capabilities of STING-BEE across four vision-language tasks: Scene Comprehension (d, e, f, i), Referring Threat Localization (a, j), Visual Grounding (c, g), and Visual Question Answering (b, h). These examples span four X-ray security datasets— STCray, SIXray, PIDray, and Compass XP — illustrating STING-BEE's robustness and adaptability to diverse X-ray imagery.}
    \label{fig_Qual2}
    \vspace{-1em}
\end{figure*}

Qualitative examples shown in \autoref{fig_Qual2} highlight STING-BEE's ability to excel across the key vision-language tasks, including scene comprehension, referring threat localization, visual grounding, and VQA, especially across diverse datasets, including STCray, SIXray, PIDray, and Compass XP, emphasizing its cross-domain generalization despite scanner variations. These results also highlight STING-BEE's robustness in handling high intra-categorical variability within threat categories, effectively addressing challenges like occlusion and object concealment in complex X-ray imagery.

\begin{figure*}[t!]
    \centering
    \includegraphics[width=1\linewidth]{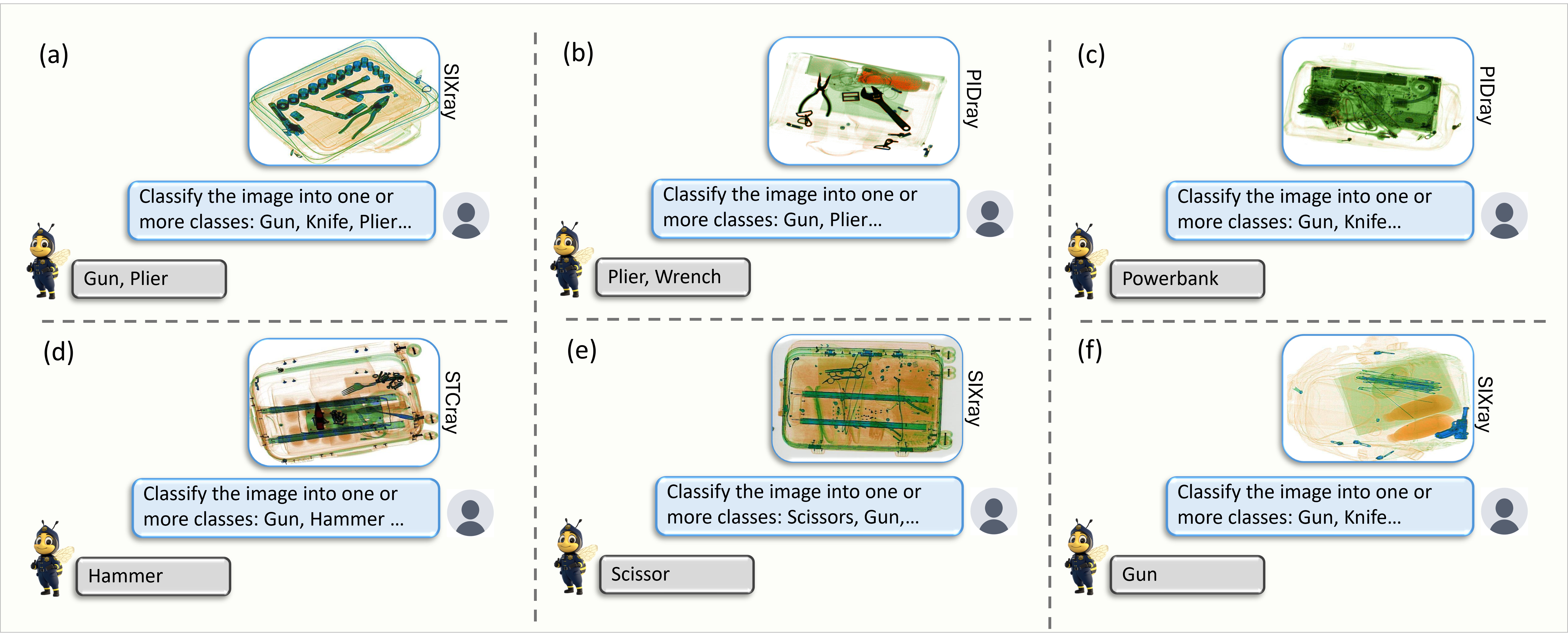} 
    \caption{Scene comprehension qualitative  examples showcasing the ability of STING-BEE to classify X-ray baggage scans into one or more threat-related classes. The images display diverse objects such as guns, pliers, wrenches, power banks, scissors, and hammers across different scenarios, highlighting the robustness of STING-BEE in understanding and categorizing threat items within X-ray imagery.}
    \label{fig_Qual3}
    \vspace{-1em}
\end{figure*}

\begin{figure*}[t!]
    \centering
    \includegraphics[width=1\linewidth]{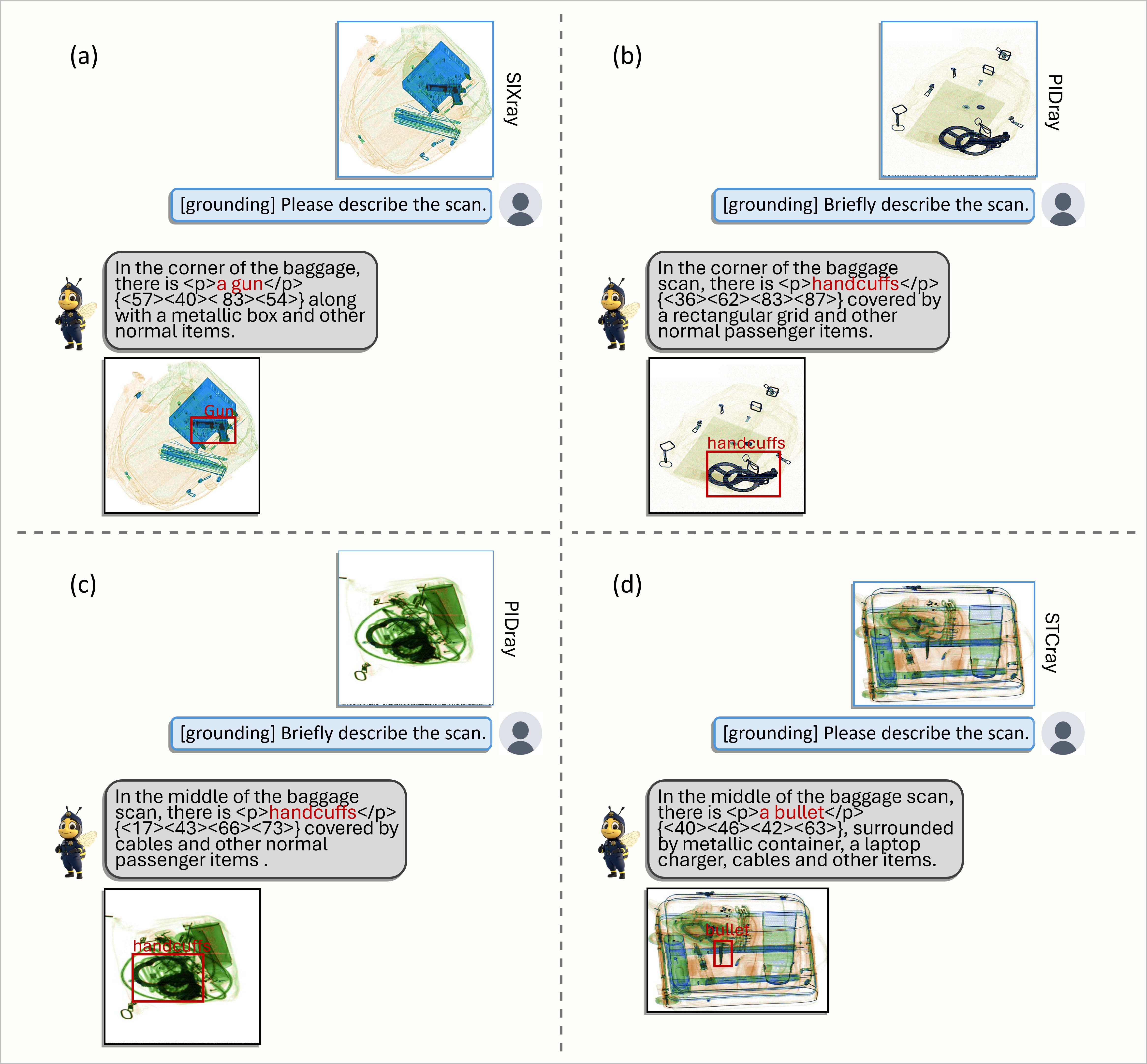} 
    \caption{Visual grounding qualitative examples demonstrating STING-BEE's ability to describe and localize specific threat items in X-ray baggage scans. The system effectively identifies and highlights objects such as guns, handcuffs, and bullets within diverse scenarios.}
    \label{fig_Qual4}
    \vspace{-1em}
\end{figure*}

\begin{figure*}[t!]
    \centering
    \includegraphics[width=1\linewidth]{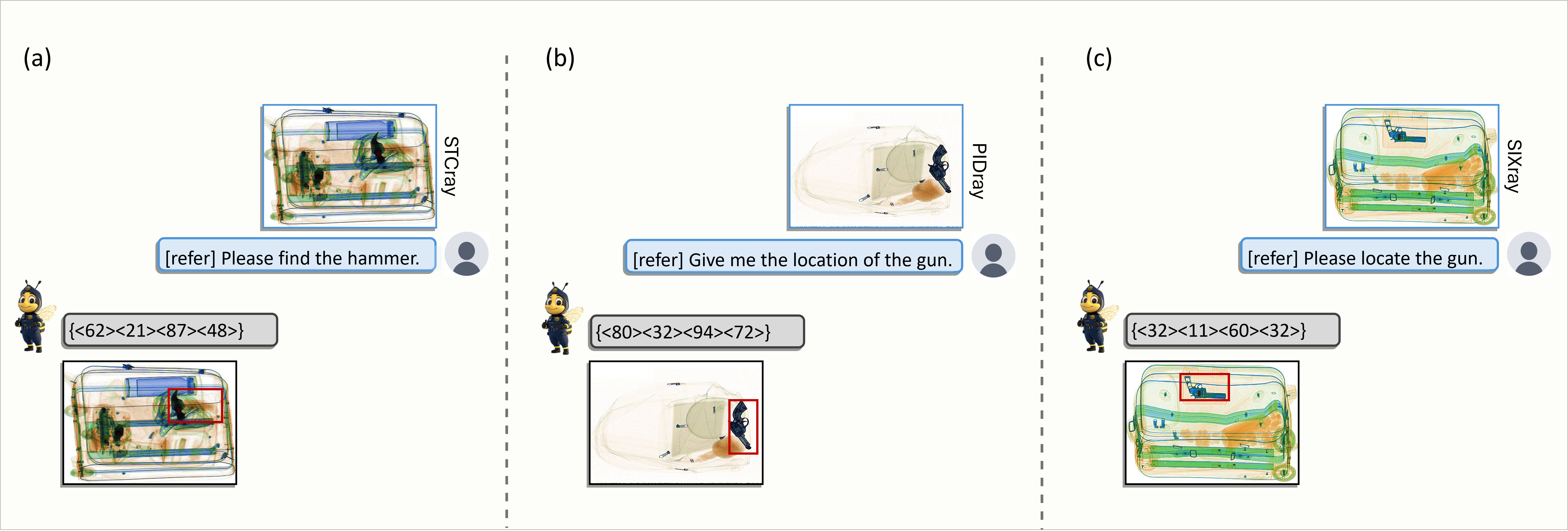} 
    \caption{Referral threat localization examples showcasing STING-BEE's precision in identifying and localizing specific threat items in X-ray baggage scans. The system demonstrates its capability to locate different contraband objects, utilizing bounding box coordinates to highlight their positions within the scans.}
    \label{fig_Qual5}
    \vspace{-1em}
\end{figure*}

\begin{figure*}[t!]
    \centering
    \includegraphics[width=1\linewidth]{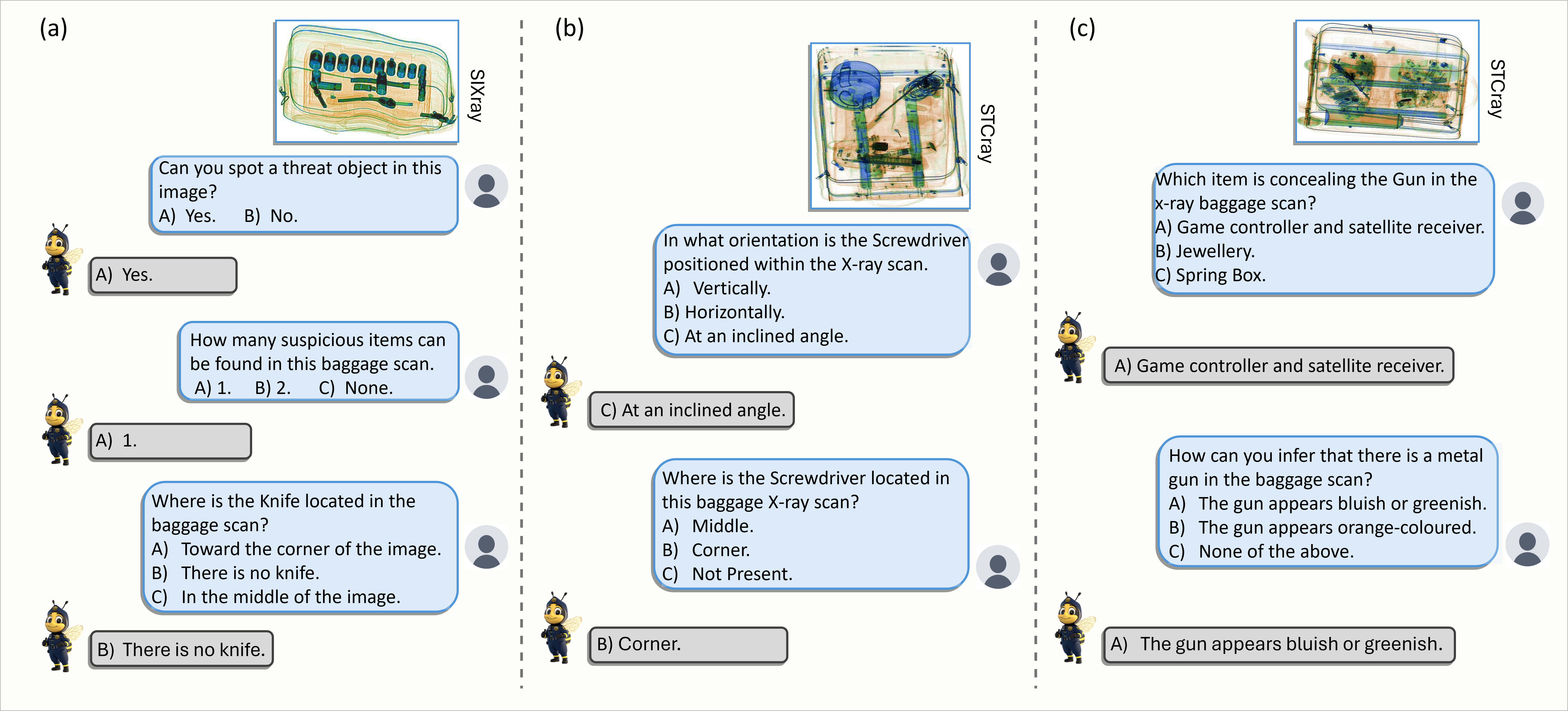} 
    \caption{Qualitative examples showcasing the capabilities of STING-BEE in Visual Question Answering (VQA) across diverse question types: (a) Instance Identification, Instance Counting, and Misleading Question resolution, (b) Instance Location and Instance Attribute recognition, and (c) Instance Interaction and Complex Visual Reasoning.}
    \label{fig_Qual6}
    \vspace{-1em}
\end{figure*}

To further illustrate these capabilities, \autoref{fig_Qual3} showcases STING-BEE's performance in scene comprehension, highlighting its ability to classify threat and non-threat items in complex baggage scans. \autoref{fig_Qual4} presents examples of visual grounding, demonstrating precise identification and localization of specific threat objects in the X-ray scans. \autoref{fig_Qual5} provides qualitative results for referring threat localization, emphasizing the model's ability to locate objects based on textual references. Finally, \autoref{fig_Qual6} showcases STING-BEE's effectiveness in handling diverse Visual Question Answering (VQA) tasks.


\appendix


\end{document}